\documentclass[manuscript,screen]{acmart}
\AtBeginDocument{%
  }

\setcopyright{acmlicensed}
\copyrightyear{2026}
\acmYear{2026}
\acmDOI{XXXXXXX.XXXXXXX}

\acmISBN{978-1-4503-XXXX-X/2026/06}

\usepackage{graphicx}
\usepackage{adjustbox}
\usepackage{subcaption}
\usepackage{booktabs}
\usepackage{amsmath}
\usepackage{multirow}
\usepackage{xcolor}

\acmJournal{health}

\begin{document}

\title{Investigating ECG Diagnosis with Ambiguous Labels using Partial Label Learning}

\author{Sana Rahmani}
\email{sana.rahmani@queensu.ca}
\affiliation{%
  \institution{Electrical and Computer Engineering Department, Queen's University}
  \city{Kingston}
  \state{Ontario}
  \country{Canada}
}
\orcid{0000-0002-2329-3446}
\author{Javad Hashemi}
\affiliation{%
  \institution{School of Computing, Queen's University}
  \city{Kingston}
  \state{Ontario}
  \country{Canada}
}
\email{j.hashemi@queensu.ca}
\orcid{0009-0005-4757-6570}
\author{Ali Etemad}
\email{ali.etemad@queensu.ca}
\orcid{0000-0001-7128-0220}
\affiliation{%
  \institution{Electrical and Computer Engineering Department, Queen's University}
  \city{Kingston}
  \state{Ontario}
  \country{Canada}
}

\renewcommand{\shortauthors}{Rahmani et al.}

\begin{abstract}
Label ambiguity is an inherent and largely unaddressed challenge in real-world electrocardiogram (ECG) diagnosis, arising from overlapping conditions and diagnostic disagreements. However, current ECG models are trained assuming clean and non-ambiguous annotations, limiting both the development and meaningful evaluation of models under real-world conditions. Although Partial Label Learning (PLL) frameworks are designed to learn from ambiguous labels, their effectiveness in medical time-series domains, ECG in particular, remains largely underexplored. We present the first systematic study of PLL methods for ECG diagnosis under both real and controlled ambiguity. First, we adapt nine PLL algorithms to multi-label ECG diagnosis under label ambiguity, and perform detailed evaluations on real clinical settings with multi-annotator diagnostic disagreements. Next, to study PLL effects on ECG in more depth under controlled settings, we introduce a diverse set of clinically motivated synthetic label ambiguities. Our experiments demonstrate that PLL methods vary substantially in robustness across ambiguity types and levels. Moreover, we observe that PLL generally outperforms standard supervised training under label ambiguity, highlighting the value of such frameworks. Through extensive analysis, we identify key limitations of current PLL approaches for clinical settings and outline future directions for developing robust and clinically aligned ambiguity-aware learning frameworks for ECG diagnosis.


\end{abstract}




\maketitle

\section{Introduction}
Electrocardiogram (ECG) signal analysis plays a pivotal role in modern cardiovascular diagnostics, offering a non-invasive and cost-effective means of monitoring the electrical activity of the heart. As cardiovascular diseases (CVDs) remain a leading cause of mortality worldwide, early and accurate detection is essential for effective intervention and treatment \cite{Johnson2025}. In recent years, advances in signal processing, machine learning, and deep learning have led to the development of automated ECG interpretation systems capable of detecting a wide range of cardiac abnormalities, including arrhythmias. These systems have demonstrated expert-level performance in tasks such as arrhythmia classification \cite{Hannun2019}, contributing to improved diagnostic efficiency and reduced clinical workload. Consequently, automated ECG analysis holds great promise for enhancing the speed, scalability, and reliability of cardiac care in both clinical and remote healthcare environments \cite{Rahmani2025}.

\begin{figure}
    \centering
    \includegraphics[width=0.9\linewidth]{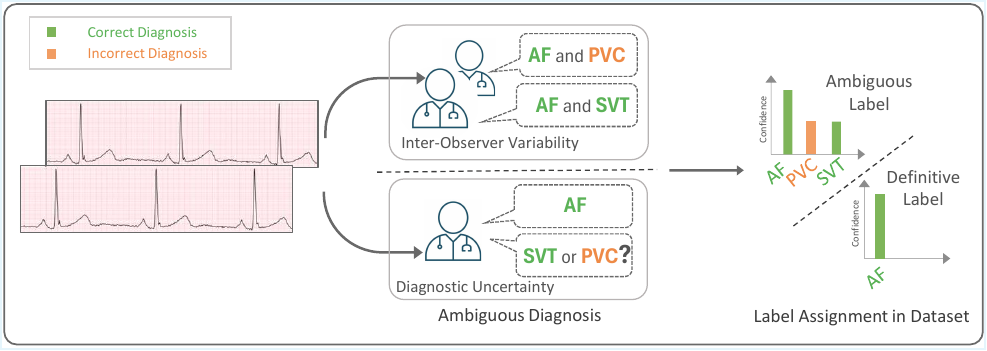}
    \caption{Ambiguity in ECG diagnosis. Inter-observer disagreement and diagnostic uncertainty can lead to multiple plausible labels for the same ECG signal. When ambiguity is preserved, the dataset reflects multiple plausible diagnoses; when suppressed, a definitive diagnosis is assigned. Selecting definitive labels is the common practice in classic supervised learning approaches.}
    \label{fig:ambiguity}
\end{figure}

One of the major challenges in real-world ECG analysis is diagnostic ambiguity, where a recording has multiple clinically plausible diagnoses \cite{Wagner2020}. This ambiguity arises in complex cases, where overlapping patterns among cardiac conditions may result in inter-observer disagreement \cite{Ribeiro2023} or diagnostic uncertainty by an expert (Fig. \ref{fig:ambiguity}). Diagnosis disagreements in ECG datasets are often resolved through consensus review \cite{Rigueira2024} or adjudication by a senior expert \cite{Ribeiro2020} to suppress ambiguity and assign a definitive diagnosis, although ambiguity may persist in clinically challenging cases. As a result, conventional supervised learning models are typically trained on labels that do not reflect the ambiguity present in real-world clinical diagnosis \cite{Grubb2020,Vandenberk2018}.

Unlike conventional supervised learning, Partial Label Learning (PLL) \cite{Tian2023} offers an effective framework that can deal with ambiguous labels. PLL methods can learn from samples annotated with a set of \textit{candidate labels} that contains the correct label set. This makes PLL particularly well-suited for biomedical and clinical settings. For example, in single-cell RNA sequencing, different PLL strategies have been used to classify cell types from ambiguous annotations, achieving performance comparable to fully supervised models while significantly reducing the need for manual curation \cite{Senoussi2024}. More recently, PLL has been extended to biosignal analysis tasks such as emotion recognition from electroencephalography (EEG), where ambiguity in self-reported emotional states is common. In these contexts, PLL methods originally developed for image-based classification have been successfully adapted to model the label ambiguity inherent in human physiological data \cite{Zhang2025}. Despite success in other biomedical fields, the use of PLL remains largely unexplored in the context of ECG diagnosis.


In this paper, for the first time, we adapt several state-of-the-art PLL methods, originally developed for image classification, to the task of cardiac diagnosis from ECG signals. We also design a comprehensive setup to evaluate them under both real-world and synthetically generated diagnostic ambiguity. First, to study PLL methods under real-world ambiguity, we use annotation disagreements among multiple clinicians in the CODE Test dataset, which, to the best of our knowledge, is the only dataset that systematically records multiple annotations per ECG sample. This provides a rare opportunity to evaluate learning methods under real clinical diagnostic ambiguity. Next, to further evaluate these methods in controlled settings, we follow the PLL literature and introduce synthetic ambiguity into the PTB-XL and Chapman datasets using six clinically motivated candidate-label generation strategies, broadly categorized into three groups: \textit{Random}, \textit{Class-Level} (i.e., Treatment Similarity, Disease Taxonomy Structure, and Cardiologist-Driven Similarity), and \textit{Instance-Level} (i.e., Model-Driven and Cardiologist-Driven Similarity). The Class-Level strategies introduce ambiguity based on inter-class similarities (e.g., clinically related conditions), while the Instance-Level strategies generate ambiguity by leveraging ECG characteristics. Together, these controlled experiments extend the primary clinical findings, enabling us to isolate the effect of ambiguity type and level under known conditions.

Our contributions in this paper are summarized as follows. \textbf{(1)} We introduce PLL to the domain of ECG diagnosis, addressing the challenge of ambiguous labels. To the best of our knowledge, this is the first work to systematically explore PLL for ECG. \textbf{(2)} We adapt and compare nine state-of-the-art PLL methods for multi-label ECG classification, analyzing their effectiveness under real clinical label ambiguity, using multi-annotator disagreement from the CODE Test dataset. \textbf{(3)} We extend the study with controlled synthetic ambiguity on PTB-XL and Chapman datasets to systematically analyze robustness across ambiguity types and levels. \textbf{(4)} To contribute to the area and enable fast reproducibility, we will release our code upon publication.

\section{Related Work}
\subsection{ECG Signal Analysis}
Recent advances in machine learning and deep learning have substantially improved automated ECG analysis, particularly in arrhythmia detection and disease classification. Most progress to date has been driven by supervised learning over large-scale annotated datasets to enable models to learn discriminative representations from ECG signals. Convolutional Neural Networks (CNNs), for example, have demonstrated strong capabilities in feature extraction from ECG waveforms \cite{Hannun2019}. Deep residual networks \cite{Li2022} have achieved cardiologist-level performance in heartbeat and arrhythmia classification, while Wang et al. \cite{Wang2019} proposed a multi-stage model combining convolutional and attention layers to capture complex waveform dependencies. To better model temporal dynamics, hybrid architectures such as CNNs combined with Long Short-Term Memory (LSTM) networks have been explored \cite{Yao2020}, with Ashhad et al. \cite{Ashhad2024} integrating LSTMs along transformer-based encoders, followed by an uncertainty-aware fusion of their outputs for an effective multi-view processing of ECG signals. More recently, transformer-based architectures such as MCTNet \cite{Zhang2024} have fused CNNs and self-attention layers to jointly capture local morphology and long-range temporal dependencies, resulting in enhanced classification performance.

Beyond supervised learning, self-supervised and semi-supervised learning approaches have been widely investigated to reduce dependence on large annotated datasets and to address challenges such as label scarcity and distribution shifts. Contrastive self-supervised learning (SSL) methods pre-train models on large collections of unlabeled ECG signals to learn generalizable representations. For example, Liu et al. \cite{Liu2025a} adopted a BYOL-inspired contrastive framework that improved arrhythmia detection performance compared to fully supervised baselines, while Liu et al. \cite{Liu2024} proposed BELL, a multi-branch variant of BYOL tailored for multi-lead ECGs, which leverages intra-lead and inter-lead objectives. Soltanieh et al. \cite{Soltanieh2023}\cite{Soltanieh2022} evaluated several SSL paradigms, including BYOL, SimCLR, and SwAV, showing that pre-trained encoders significantly improve generalization to both in-distribution and out-of-distribution tasks, with SwAV offering the strongest performance. Semi-supervised methods have also been introduced, such as CPSS \cite{Shi2024}, which combines consistency regularization with pseudo-labeling by assigning positive pseudo-labels to highly confident predictions and negative pseudo-labels to low-confidence categories. By enforcing prediction consistency across weak and strong augmentations, CPSS improves learning from limited labeled data.

Despite these advances in supervised, semi-supervised, and self-supervised ECG analysis, PLL remains largely unexplored in this domain. Unlike semi-supervised learning, which assumes access to both labeled and unlabeled data, PLL assumes that each training instance is associated with a candidate label set containing the true label set but also false positive ones. This paradigm naturally captures real-world annotation scenarios in ECG diagnosis, where diagnostic uncertainty or inter-observer disagreement may result in ambiguous labels. As such, PLL provides a complementary direction to existing weak supervision strategies, with the potential to improve robustness and generalization in automated ECG interpretation. This motivates our adaptation and systematic study of PLL methods for ECG signal analysis.

\subsection{Partial Label Learning}
PLL has emerged as a powerful framework for addressing label ambiguity, where each training instance is associated with a set of candidate labels containing the unknown true label. PLL methods are broadly categorized into two major families: \textbf{(1) Average-based methods} treat all candidate labels equally, training the model by averaging supervision signals across all labels in the candidate set. These methods are computationally efficient but often suffer from false positives due to the absence of mechanisms to disambiguate incorrect candidate labels. A representative method is Deep Naive Partial Label (DNPL) \cite{Seo2021}, which employs a custom loss function that assigns higher probability mass to candidate labels without explicitly resolving the ambiguity within the candidate label set. \textbf{(2) Identification-based methods}, on the other hand, aim to explicitly infer the ground-truth label from the candidate set by assigning confidence scores to each label and jointly optimizing the model parameters and the label assignments. This class of methods has gained increasing attention for its superior robustness to label ambiguity. For instance, Kim et al. \cite{Kim2022} propose filtering out samples with high fluctuations on loss to mitigate ambiguity, although such strategies may inadvertently suppress minority class signals. To overcome class imbalance, the proposed method in \cite{Wang2024} introduces an optimization-based disambiguation that accounts for inter-class confidence differences. Similarly, Wang et al. \cite{Wang2022b} utilize Sinkhorn distance regularization \cite{Cuturi2013} to balance label confidence across classes, thereby improving learning in underrepresented categories.

Several recent methods further enhance label disambiguation by leveraging model parameters. Zhang et al. \cite{Zhang2021} propose a gradient-based label weighting mechanism, where gradients of model outputs guide candidate label importance. Prototype-based approaches \cite{Wang2022a} estimate class centroids in embedding space to assign labels via instance-prototype similarity, which are further used to construct supervised contrastive pairs. Similarly, Wu et al. \cite{Wu2020} incorporate manifold consistency regularization to enforce label smoothness in the feature space, improving label consistency among similar instances. Beyond architectural strategies, loss function modifications have also proven effective in improving PLL performance. Curriculum learning methods such as \cite{Durand2019} introduce a training schedule based on sample difficulty, normalizing the binary cross-entropy (BCE) loss according to candidate label sizes. Other methods explore learning from both candidate and non-candidate labels. Wen et al. \cite{Wen2021} apply distinct loss functions to each set: a modified sigmoid loss for candidate labels and a standard sigmoid loss for non-candidate ones. Similarly, the method in \cite{Wu2022} employs consistency regularization over augmented inputs to enforce stable predictions under input perturbations, improving generalization under ambiguous supervision. Collectively, PLL methods use architectural, optimization, and loss-based strategies to address label ambiguity. While average-based approaches offer computational simplicity, identification-based methods provide stronger robustness by estimating and correcting label confidences, making them suitable for real-world applications such as medical diagnosis.

\subsection{PLL in Other Medical Domains}
PLL is increasingly used in biomedical domains by enabling models to learn from ambiguous or weakly supervised labels while reducing the dependence on large amounts of expert-annotated data. A prominent application area is single-cell RNA sequencing, where both hierarchical and non-hierarchical PLL strategies have been leveraged to identify cell type, showing accurate classification of transcriptomic profiles is feasible with substantially fewer annotations than required by fully supervised methods \cite{Senoussi2024}. In blood cell classification, Feng et al. \cite{Feng2024} proposed a PLL-based strategy that utilizes weakly annotated data in combination with morphological domain knowledge to construct candidate label sets. This method not only improved classification accuracy but also enhanced model interpretability in identifying diverse blood cell attributes. Similarly, in whole slide image (WSI) analysis, PLL has been applied to address the challenge of incomplete labeling. The proposed method in \cite{Matsuo2024} incorporates partial label proportions like subtype ratios among tumor categories, to accurately classify image patches as tumor or non-tumor regions. 

Recent work has also explored PLL in the analysis of human biosignals, particularly EEG for emotion recognition. Zhang et al. \cite{Zhang2025} adapted image-based PLL methods to learn from EEG signals, successfully handling label ambiguity during training. To further improve generalization under noisy supervision, Li et al. \cite{Li2024a} introduced a generative contrastive loss with feature separability constraints, resulting in more robust models for EEG classification. While these advances illustrate the versatility of PLL across biomedical settings, its application to other physiological signals like ECG, photoplethysmography (PPG), and electromyography (EMG), remains limited. Extending PLL methodologies to these domains presents an important opportunity for developing robust diagnostic systems capable of operating under real-world conditions where label ambiguity is prevalent. In this work, we focus on bridging this gap by exploring the first systematic application of PLL to ECG diagnosis.

\section{Methods}
\subsection{Preliminaries}
Let $x \in \mathbb{R}^d$ denote an input ECG signal with length $d$, and $\mathcal{Y} = \{1, \dots, C\}$ represent the set of $C$ possible diagnostic labels. In the PLL framework, each data instance $x_i$ is associated with a candidate label set $Y_i \subseteq \mathcal{Y}$, which contains the unknown ground-truth label set $\tilde{Y}_i \subseteq Y_i$. 
In multi-label settings, the true label set $\tilde{Y}_i$ may include one or more correct classes, while the remaining classes in $Y_i$ are the label ambiguities introduced to $x_i$. 
Given a training dataset $\mathcal{D}_t = \{(x_i, Y_i)\}_{i=1}^{n}$, the goal is to learn a multi-label classification model $\mathbf{f}: \mathbb{R}^d \rightarrow [0,1]^C$ that estimates a probability distribution over the label set $\mathcal{Y}$ for each input $x_i$. The model should minimize the empirical risk over the ambiguous supervision, where supervision is defined over candidate label sets:
\begin{equation}
    \min_{\theta}(\frac{1}{n} \sum_{i=1}^{n} \mathcal{L}(\mathbf{f}(x_i; \theta), Y_i)).
\end{equation}
Here, $\mathbf{f}(x_i; \theta)$ is the output of the model parameterized by $\theta$, and $\mathcal{L}(\cdot)$ is a PLL-aware loss function that accounts for the ambiguity in $Y_i$. 

In this work, we adapt and comprehensively analyze nine state-of-the-art PLL algorithms for multi-label ECG classification. These models were selected based on the diverse strategies they employ to address the PLL problem. For example, some approaches focus on contrastive learning to enhance training quality (e.g., Partial Label Learning with Contrastive Label Disambiguation \cite{Wang2022a}), while others leverage negative supervision to ensure that all available supervision information is utilized (e.g., Consistency Regularization \cite{Wu2022} and Leveraged Weighted \cite{Wen2021}). Additionally, certain methods address multiple challenges inherent in real-world PLL scenarios, such as long-tailed label distributions in the training data (e.g., COrrection ModificatIon balanCe \cite{Zhang2023a}). To provide clarity and easier comparison across PLL methods, Table \ref{tab:LossFuncComp} summarizes loss function and disambiguation mechanism, while Table \ref{tab:pll_comparison} compares their components, including integration flexibility, use of semantic information, use of negative supervision, and handling long-tailed distributions. 
The selected models, described in detail below, were originally developed for image-based tasks and have been adapted here to accommodate the temporal and multi-label nature of ECG signals. 
Details of our adaptations of PLL baselines are presented within each section below. 
Each method is evaluated under controlled ambiguity conditions to assess its effectiveness in learning from candidate label supervision.

In Fig. \ref{fig:framework}, we illustrate the overall framework of a PLL-based ECG diagnosis model. A typical PLL framework consists of two key components: the \textit{Disambiguation} module and the PLL loss function ($\mathcal{L}_{PLL}$). The model is trained on a dataset containing both ambiguous and non-ambiguous samples, which necessitates the use of an ambiguity-aware loss. During training, each input ECG and its candidate label set are fed into the encoder–classifier pipeline to produce predictions, which are optimized via backpropagation. In parallel, the disambiguation module refines the candidate label set by reweighting label confidences based on multiple sources of evidence, such as model prediction scores, gradient information, or label co-occurrence statistics. By iteratively combining disambiguation with supervised optimization, the framework progressively reduces annotation ambiguity and improves classification accuracy.
\begin{figure}
    \centering
    \includegraphics[width=\linewidth]{img/figures/framework_pll.pdf}
    \caption{Overall framework of a PLL-based ECG diagnosis model. The framework integrates an encoder–classifier with a disambiguation module, where candidate label sets are iteratively refined using information such as prediction confidence, and gradient information, while training is guided by an ambiguity-aware loss.}
    \label{fig:framework}
\end{figure}

\begin{table*}[t]
\centering
\small
\caption{Comparison of loss functions and disambiguation strategies of different PLL models.}
\label{tab:LossFuncComp}
\renewcommand{\arraystretch}{1.5} 
\begin{adjustbox}{max width=0.9\textwidth}
\begin{tabular}{l p{0.48\textwidth} p{0.42\textwidth}}
\toprule
\textbf{PLL Model} & \textbf{Loss Function} & \textbf{Disambiguation Function} \\
\midrule
\midrule
\textbf{DNPL \cite{Seo2021}} & 
$\mathcal{L}_{\text{DNPL}}=-\frac{1}{n}\sum_{i=1}^{n} 
\log\!\left(\langle\sigma(\mathbf{f}(x_i;\theta)),Y_i\rangle\right)$ 
& No disambiguation \vspace{5mm}\\

\textbf{PRODEN \cite{Lv2020}} & 
$\mathcal{L}_{\text{PRODEN}}=\sum_{i=1}^{n}\min_{y_j\in Y_i}
L(f_j(x_i;\theta),y_j)$ 
& $w_{i,j}= \dfrac{f_j(x_i,\theta)}{\sum_{y_{z} \in Y_i} f_z(x_i,\theta)}$ \vspace{5mm}\\

\textbf{CAVL \cite{Zhang2021}} & Supervised cross-entropy loss &
$w_{i,j}=\left| f_j(x_i;\theta) - 1 \right| f_j(x_i;\theta)$ \vspace{5mm}\\

\textbf{LW \cite{Wen2021}} &
$\mathcal{L}_{LW}=\sum_{y_j \in Y_i} w_{i,j} L(f_j(x_i))
+ \beta \cdot \sum_{y_j \notin Y_i} w_{i,j} L(-f_j(x_i))$
& 
$\begin{aligned}
w_{i,j} &= 
\begin{cases}
\dfrac{e^{f_j(x_i;\theta)}}{\sum_{y_k\in Y_i} e^{f_k(x_i;\theta)}} & y_j \in Y_i \\[8pt]
\dfrac{e^{f_j(x_i;\theta)}}{\sum_{y_k\notin Y_i} e^{f_k(x_i;\theta)}} & y_j \notin Y_i
\end{cases}
\end{aligned}$ \vspace{5mm}\\

\textbf{CR \cite{Wu2022}} & 
$\mathcal{L}_{CR}=-\sum_{y_j\notin Y_i}\log (1-f_j(x_i,\theta)) 
+ \gamma_t\!\sum_{\tilde{x}\in \widetilde{X}_i}\!KL(p||f(\tilde{x}))$
& 
$p_k=\dfrac{\left(\prod_{\tilde{x}\in \widetilde{X}_i} f_{k}(\tilde{x})\right)^{\tfrac{1}{|\widetilde{X}_i|}}}
{\sum_{y_j \in Y_i}\left(\prod_{\tilde{x}\in \widetilde{X}_i} f_{j}(\tilde{x})\right)^{\tfrac{1}{|\widetilde{X}_i|}}}$ \vspace{5mm}\\

\textbf{PICO \cite{Wang2022a}} & 
$\mathcal{L}_{\text{PICO}} = \sum_{i=1}^{n} 
\big( l_{\text{cl}}(x_i, Y_i) + \lambda l_{\text{co}}(x_i)\big)$ 
& 
$\begin{aligned}
Y_i &\leftarrow \alpha Y_i + (1-\alpha)\mathbf{w}_i, \\
w_{i,j} &= \begin{cases}
1 & \text{if } j=\arg\max_{y_j \in Y_i}\mathbf{q}_i^\top \boldsymbol{\mu}_j \\
0 & \text{otherwise}
\end{cases}
\end{aligned}$ \vspace{5mm}\\

\textbf{SST \cite{Chen2022a}} & 
$\mathcal{L}_{SST} = L_{cls} + \lambda_1 L_{ist} + \lambda_2 L_{cst}$ 
& 
$\begin{aligned}
\tilde{y}_k &= \mathbf{1}\!\left[\Big(\sum_{y_j \in Y_i} p^{ist}_{k,j} y_j\Big) \geq \theta^{ist} \right] \\
\widehat{y}_j^i &= \mathbf{1}\!\left[\Big(\tfrac{1}{|\mathcal{D}_j|}\sum_{x_z \in \mathcal{D}_j} s^j_{i,z}\Big) \geq \theta^{cst} \right]
\end{aligned}$ \vspace{5mm}\\

\textbf{HST \cite{Chen2024a}} & 
$\mathcal{L}_{HST} = L_{cls} + \lambda_1 L_{ist} + \lambda_2 L_{cst} 
+ \lambda_3\!\left(L_{ist}^{dtl} + L_{cst}^{dtl}\right)$
& 
(same disambiguation as SST) \vspace{5mm}\\

\textbf{COMIC \cite{Zhang2023a}} & 
$\mathcal{L}_{COMIC} = \lambda_m L_m + \lambda_b L_b + \lambda_c L_c$
& 
$\widehat{y}_{j}=\begin{cases}
1 & \text{if } p_{c}>\max \{\tau, P_{j}\},\ y_{j}=1,\\
0 & \text{otherwise}
\end{cases}$ \\
\bottomrule
\end{tabular}
\end{adjustbox}
\end{table*}

\begin{table*}[t]
\centering
\small
\caption{Overview of different PLL methods. \checkmark and $\times$ respectively indicate inclusion and exclusion of a component in the method. }
\label{tab:pll_comparison}
\resizebox{\textwidth}{!}{
\begin{tabular}{l c c c c c p{3cm} p{3.2cm} p{3cm}}
\toprule
\textbf{Method} & \textbf{Integration Flex.} & \textbf{Neg. Sup.} & \textbf{Semantic Info} & \textbf{Long-tail Handel} & \textbf{Aug. Rep.} & \textbf{Loss Function Design} & \textbf{Disambiguation Strategy} & \textbf{Strengths / Intended Scenarios} \\
\midrule
\midrule
\textbf{DNPL\cite{Seo2021}}   & High     & $\times$ & $\times$ & $\times$ & $\times$ & Naive PLL loss (inner product with candidate set) & None (naive distribution alignment) & Simple, low complexity \\
\textbf{PRODEN \cite{Lv2020}} & High     & $\times$ & $\times$ & $\times$ & $\times$ & Refinement-based CE loss & Prediction-based distribution refinement & Lightweight training with disambiguation \\
\textbf{CAVL \cite{Zhang2021}}   & High     & $\times$ & $\times$ & $\times$ & $\times$ & Refinement-based CE loss & CAV-based refinement & Low-cost disambiguation via class activation vectors \\
\textbf{LW \cite{Wen2021}}     & High     & \checkmark & $\times$ & $\times$ & $\times$ & Weighted binary loss on pos./neg. labels & Implicit disambiguation via weighted loss & Leverages all available supervision \\
\textbf{CR \cite{Wu2022}}     & Moderate & \checkmark & $\times$ & $\times$ & \checkmark & Log-likelihood on neg. labels + KL-based consistency loss & Progressive refinement from model predictions & Encourages consistency under augmentations \\
\textbf{PICO \cite{Wang2022a}}   & Moderate & $\times$ & \checkmark & $\times$ & \checkmark & Contrastive PLL + log-based classification loss & Contrastive disambiguation & Robust to noisy candidate sets via representation learning \\
\textbf{SST \cite{Chen2022a}}    & Low      & $\times$ & \checkmark & $\times$ & $\times$ & Binary CE + semantic module-specific loss & Per-module refinement with adaptive thresholds & Exploits inter-class and inter-sample dependencies \\
\textbf{HST \cite{Chen2024a}}    & Low      & $\times$ & \checkmark & $\times$ & $\times$ & SST loss + binary CE on thresholds & Threshold-based refinement with automation & Efficient, automated threshold version of SST \\
\textbf{COMIC \cite{Zhang2023a}}  & Moderate & $\times$ & \checkmark & \checkmark & $\times$ & MFM long-tail loss + balance loss + refinement-based classification & Class-distribution refinement with multi-model balancing & Excels under long-tail distributions and high ambiguity \\
\bottomrule
\end{tabular}}
\end{table*}

\subsection{Deep Naive Partial Label}
The DNPL model \cite{Seo2021} performs partial label learning without explicit disambiguation. Instead of attempting to identify the ground-truth label within the candidate label set, DNPL optimizes a naive loss function that encourages the model's predictions to align with the entire candidate label distribution. The DNPL loss is defined as:
\begin{equation}
    \mathcal{L}_{\text{DNPL}} = -\frac{1}{n} \sum_{i=1}^{n} \log \left( \left\langle \sigma(\mathbf{f}(x_i;\theta)), Y_i \right\rangle \right).
\end{equation}
Here, $\left\langle \cdot, \cdot \right\rangle$ denotes the inner product between the predicted probability vector and the binary candidate label vector, and $\sigma$ denotes the softmax activation function. This formulation encourages the model to place higher probability on the candidate labels, effectively learning under weak supervision. Since the original DNPL framework was designed for image data, we adapt the architecture to ECG signals by employing a ResNet18 backbone tailored for time-series signals, while retaining the original loss function $\mathcal{L}_{\text{DNPL}}$ to address the PLL setting. While DNPL does not attempt to resolve ambiguity by estimating true labels, it provides a simple and effective baseline that leverages the structure of candidate label supervision directly in the loss function.

\subsection{Progressive Identification} 
The proposed method in PROgressive iDENtification (PRODEN) \cite{Lv2020} defines a risk estimator compatible with stochastic optimization for the PLL problem. Accordingly, the PLL loss is defined for classification as the minimum loss among the candidate label set as follows
\begin{equation}
    \mathcal{L}_{\text{PRODEN}} = \sum_{i=1}^{n} \min_{y_j\in Y_i} L \left( f_j(x_i;\theta),y_j  \right) ,
    \label{EQ:PRODEN loss}
\end{equation}
where $L$ is the cross entropy loss, and $f_j$ denotes the model prediction on the $j^{\text{th}}$ label. 

Unlike the DNPL, PRODEN performs disambiguation on the candidate labels parallel to training and tries to converge the model prediction with the disambiguated label set. Accordingly, the $min$ operator in Eq. \ref{EQ:PRODEN loss} is replaced by dynamic weights $w_{i,j}$ over the candidate label set which demonstrates the model confidence on each label being the true label.
To this end, we initialize the weights uniformly among the candidate set (i.e., $w_{i,j}=\dfrac{1}{|Y_i|}$), and further disambiguate this weight in each epoch using model predictions to encourage the possible labels with more weight, as follows
\begin{equation}
    w_{i,j}= \dfrac{f_j(x_i,\theta)}{\sum_{y_{z} \in Y_i} f_z(x_i,\theta)}.
    \label{eq:proden_dis}
\end{equation}
Similar to the DNPL framework, we adapt the PRODEN model for disease classification by employing a ResNet18 backbone designed for time-series. In this adaptation, we retain the original PRODEN loss function $\mathcal{L}_{\text{PRODEN}}$, and apply the disambiguation procedure defined in Eq.~\ref{eq:proden_dis} at each training iteration within our PLL setting.

\subsection{Class Activation Value Learning }
The Class Activation Value Learning (CAVL) \cite{Zhang2021} method progressively identifies the true label from a candidate label set by estimating the importance of each label based on learned class-specific activation information. CAVL offers an intuitive and gradient-sensitive approach to disambiguate labels without relying on external heuristics. Inspired by Gradient-weighted Class Activation Mapping (Grad-CAM) \cite{Selvaraju2020}, CAVL computes an importance weight, known as the Class Activation Value (CAV), for each candidate label. This score is derived from the gradient information of the last layer and class-specific output of the model. The CAV for class $j$ and instance $x_i$ is computed as:
\begin{align}
    w_{i,j} &= \left| \frac{\partial \left( -\log \sigma_j(f(x_i;\theta)) \right)}{\partial f_j(x_i;\theta)} \right| \sigma_j(f(x_i;\theta)) \notag \\ 
   &= \left| \sigma_j(f(x_i;\theta)) - 1 \right| \sigma_j(f(x_i;\theta)).
    \label{EQ:CAVL weight}
\end{align}

Since the raw model output $f(x_i; \theta)$ captures richer discriminative information than the class probability $\sigma(f(x_i; \theta))$, the softmax activation is removed in practice. This leads to the simplified CAV formulation:
\begin{equation}
    w_{i,j}=\left| f_j(x_i;\theta) - 1 \right| f_j(x_i;\theta).
\end{equation}
To disambiguate the candidate label set $Y_i$, CAVL selects the label with higher CAV importance score among the candidates. 
As the original CAVL architecture was proposed for image classification, we adapt its disambiguation strategy to disease classification by employing a ResNet18 backbone designed for time-series signals. In this setting, the disambiguated labels $\widehat{y}_i$ are subsequently used to compute the supervised cross-entropy loss during model training.

\subsection{Leveraged Weighted}
Most existing PLL algorithms focus exclusively on candidate labels, often ignoring the complementary information provided by non-candidate classes. To this end, Leveraged Weighted (LW) \cite{Wen2021} incorporates both candidate and non-candidate labels into the training process. The core idea is to apply weighted supervision over all classes, where candidate labels are promoted and non-candidate labels are explicitly discouraged. To this end, LW combines candidate and non-candidate labels based on binary loss $L$ such as sigmoid for a classification loss as follows: 
\begin{equation}
    \mathcal{L}_{LW}=\sum_{y_j \in Y_i} w_{i,j} L\left(f_j(x_i)\right)+\beta \cdot \sum_{y_j \notin Y_i} w_{i,j} L\left(-f_j(x_i)\right).
\end{equation}
Here, $L\left(f_j(x_i)\right)$ encourages the model for predictions within the candidate label set and the slightly different component $L\left(-f_j(x_i)\right)$ discourages predictions outside the candidate label set. This combination of prediction loss among candidate and non-candidate labels is controlled with a parameter $\beta$. 

To obtain the weight $w_{i,j}$, we normalize the model predictions for both the candidate and non-candidate labels, where:
\begin{equation}
    w_{i,j} = \left\{ 
    \begin{array}{rl}
        \dfrac{\text{exp}(f_j(x_i;\theta))}{\sum_{y_k\in Y_i}\text{exp}(f_k(x_i;\theta))}, & y_j \in Y_i \\[12pt]
        \dfrac{\text{exp}(f_j(x_i;\theta))}{\sum_{y_k\notin Y_i}\text{exp}(f_k(x_i;\theta))}, & y_j \notin Y_i
    \end{array}
    \right.
\end{equation}
Progressively, this weight is updated to encourage disambiguation of the candidate label set, where a higher weight $w_{i,j}$ would be given to labels which are more likely to be positive, and ambiguity in labels is further decreased. The disambiguation strategy proposed in LW is applied to obtain the LW loss $\mathcal{L}_{LW}$, which is used for training the backbone ResNet18 model on ambiguous ECG signals.

\subsection{Consistency Regularization}
A method similar to LW is adopted in the Consistency Regularization (CR) \cite{Wu2022} approach, where non-candidate labels are used in the training objective. CR applies a negative log-likelihood loss over non-candidate labels to exploit the certainty of negative supervision. Additionally, CR introduces a consistency regularization term that enforces the model’s predictions to remain stable across different augmentations of the same input, ensuring robust learning from candidate labels.
To achieve this, a regularization loss is defined between the model's predictions on a set of augmented inputs $\widetilde{X}_i$ and a conformal label distribution $p$. The conformal label distribution is a transformed version of the model's output probabilities, representing the model's confidence across candidate classes. It is initialized with a uniform distribution and progressively optimized during training based on the model’s predictions as follows:
\begin{equation}
p_k=\frac{\left(\prod_{\tilde{x}\in \widetilde{X}_i} f_{k}(\tilde{x})\right)^{\frac{1}{|\widetilde{X}_i|}}}{\sum_{y_j \in Y_i}\left(\prod_{\tilde{x}\in \widetilde{X}_i} f_{j}(\tilde{x})\right)^{\frac{1}{|\widetilde{X}_i|}}} .
\end{equation}
Here, $f_j(\tilde{x})$ denotes the model's prediction for class $j$ on the augmented sample $\tilde{x}$.
Given an input ECG signal $x_i$, we generate a set of augmented views $\widetilde{X}_i = \mathcal{A}(x_i)$ using a data augmentation function $\mathcal{A}$. A common practice for augmenting ECG signals \cite{Soltanieh2023} is to apply Additive White Gaussian Noise (AWGN), defined as $\mathcal{A}(x_i) = x_i + \mathcal{N}(\mu, \sigma)$. This approach enables the generation of both weak and strong augmentations, following the methodology in \cite{Zhang2025}.

The final CR loss at training epoch $t$ is defined as:
\begin{equation}
     \mathcal{L}_{CR}=-\sum_{y_j\notin Y_i}\log (1-f_j(x_i,\theta)) +\gamma_t\sum_{\tilde{x}\in \widetilde{X}_i}KL(p||f(\tilde{x})).
\end{equation}
The first term penalizes high-confidence predictions on non-candidate labels, preventing predictions on labels not in the candidate label set $Y_i$. The second term applies consistency regularization by minimizing the Kullback–Leibler (KL) divergence between the model predictions on augmented inputs and the conformal label distribution $p$.

To balance the contribution of the two loss components during training, a dynamic weighting factor $\gamma_t$ is introduced:
\begin{equation}
    \gamma_t=\min(\dfrac{t}{T}\lambda,\lambda)
\end{equation}
Here, $\gamma_t$ gradually increases over training epochs $t$ until it reaches a maximum value $\lambda$ at epoch $T$. This annealing strategy emphasizes learning from non-candidate labels in early stages and progressively increases the influence of model prediction consistency, thereby encouraging the model to make confident predictions. We adapt the CR framework to disease classification from ECG signals by employing a ResNet18 backbone tailored for time-series. We then apply the resulting CR loss $\mathcal{L}_{CR}$ to train the model on ambiguous ECG data.

\subsection{Partial Label Learning with Contrastive Label Disambiguation}
The Partial Label Learning with Contrastive Label Disambiguation (PICO) \cite{Wang2022a} framework integrates contrastive learning with a prototype-based label disambiguation strategy to enhance representation learning under partial label supervision. Its core idea is to extract class-consistent feature representations and leverage them to assign more accurate soft labels. To this end, PICO defines class prototypes as representative embeddings of each class and updates them progressively. 

In supervised contrastive learning, true labels are used to define positive and negative pairs. However, this is infeasible in PLL setting, where the ground-truth label is unknown and the candidate label set may contain false positives. Unsupervised contrastive learning, on the other hand, constructs positive pairs using augmented views of the same input. PICO adopts this strategy by generating two views of an input sample $x$ using data augmentations $\mathcal{A}_q$ and $\mathcal{A}_k$, producing query and key embeddings: $\mathbf{q} = g(\mathcal{A}_q(x))$ and $\mathbf{k} = g'(\mathcal{A}_k(x))$, where $g$ and $g'$ are encoder networks.

Key embeddings are stored in a queue to form an embedding pool: $A = B_q \cup B_k \cup \text{queue}$, and the positive set is defined as $p(x) = \left\{ x' \in A(x) \,\middle|\, f(\mathcal{A}_q(x')) = f(\mathcal{A}_q(x)) \right\}$, 
where $A(x) = A \setminus \{ \mathbf{q} \}$. The contrastive loss is then defined by contrasting the query embedding against all embeddings in the embedding pool:
\begin{equation}
    l_{\text{co}}(x) = -\frac{1}{|p(x)|} \sum_{x'' \in p(x)} \log \frac{\exp(\mathbf{q}^\top x'' / \tau)}{\sum_{x' \in A(x)} \exp(\mathbf{q}^\top x' / \tau)},
\end{equation}
where $\tau$ is a temperature scaling factor.

In addition to contrastive learning, PICO includes a classification loss to guide predictions toward the candidate labels:
\begin{equation}
    l_{\text{cl}}(x, Y) = \sum_{y_j \in Y} -y_j \log(f_j(x; \theta)).
\end{equation}

The total loss of PICO combines both components:
\begin{equation}
    \mathcal{L}_{\text{PICO}} = \sum_{i=1}^{n} \left( l_{\text{cl}}(x_i, Y_i) + \lambda l_{\text{co}}(x_i)   \right),
\end{equation}
where $\lambda$ balances the contribution of the classification loss.

To support label disambiguation, class prototypes $\boldsymbol{\mu}_j$ are updated as the mean of embeddings assigned to class $j$ by the classifier $\boldsymbol{\mu}_j \leftarrow \text{avg} \left\{ \mathbf{q}_i \,\middle|\, j = \arg\max f(\mathcal{A}_q(x_i)) \right\}$.

Using these prototypes, each candidate label set $Y_i$ is refined using a moving average update:
\begin{align}
Y_i &\leftarrow \alpha Y_i + (1 - \alpha) \mathbf{w}_i, \\
w_{i,j} &= 
\begin{cases}
1 & \text{if } j = \arg\max_{y_j \in Y_i} \mathbf{q}_i^\top \boldsymbol{\mu}_j \\
0 & \text{otherwise},
\end{cases}
\label{eq:picodis}
\end{align}
where $\alpha \in [0,1]$ controls the smoothing rate. This update gradually aligns the candidate labels with the most semantically similar prototypes, facilitating more confident disambiguation. We adapt the PiCO framework to disease classification from ECG by employing a ResNet18 backbone. During training, we apply the disambiguation process defined in Eq.~\ref{eq:picodis} together with the PiCO loss $\mathcal{L}_{\text{PICO}}$ to train the model.

\subsection{Structured Semantic Transfer}
Structured Semantic Transfer (SST) \cite{Chen2022a} is a PLL framework that enhances label disambiguation by exploiting structured semantic relationships between classes. Unlike conventional PLL methods that rely solely on candidate label sets, SST explicitly models inter-class correlations and transfers semantic structures to guide learning under ambiguity, thereby improving prediction, consistency, and robustness.

The SST framework begins by encoding an input ECG signal $x_i$ into a feature representation $\phi_i$ using a backbone neural network. In our adaptation, we employ a ResNet18 backbone to extract ECG-specific representations, which are then passed to subsequent modules for further processing. These features are further refined through a semantic decoupling \cite{Chen2019} module that generates class-specific, semantic-aware representations $\tilde{\phi}_{i,j}$ for each class $j$. In the original SST design for image classification, class semantics are derived from pre-trained word embeddings. However, to ensure a fair comparison of partial label learning capabilities across all evaluated methods, we restrict the use of external resources such as pre-trained language models or ontology-based embeddings. Instead, we replace the original class semantic vectors with prototype embeddings computed as the centroid of ECG signals belonging to each class. This design choice isolates the effect of the semantic decoupling architecture itself, enabling us to assess its contribution under identical information constraints for all models. Although centroid-based embeddings may lack the external semantic richness of word embeddings, they provide a dataset-derived proxy for class semantics.

In this module, the feature representation $\phi_i$ is fused with the semantic embedding $e_j$ via low-rank bilinear pooling as follows:
\begin{equation}
    \tilde{\phi}_{i,j}=\mathbf{P}^{T}\left(\tanh \left(\left(\mathbf{U}^{T} \phi_i\right) \odot\left(\mathbf{V}^{T} e_j\right)\right)\right)+\mathbf{b},
\end{equation}
where $\tanh$ is the hyperbolic tangent function, $\odot$ denotes element-wise multiplication, and $\mathbf{P}$, $\mathbf{U}$, $\mathbf{V}$, and $\mathbf{b}$ are learnable parameters. The resulting semantic-aware representations are then passed through a sigmoid-activated classifier to produce the output probabilities $\mathbf{p}_i$.

Given that certain classes are semantically correlated, the SST framework leverages inter-class dependencies to enrich class-specific representations. In ECG analysis, for instance, some conditions frequently co-occur, and capturing these relationships can improve label disambiguation. To model such correlations, we construct an adjacency matrix $A \in [0,1]^{C\times C}$ that encodes class co-occurrence patterns. Specifically, in our adaptation, $A$ is derived directly from the partial labels in the ECG training dataset, ensuring that the graph captures clinically meaningful dependencies as closely as possible without relying on ground-truth labels. Each entry $A_{j,k}$ is defined as the probability of observing class $k$ given class $j$, computed as the number of times classes $j$ and $k$ co-occur divided by the total number of occurrences of class $j$. This matrix defines a weighted graph over the class space, where edges represent co-occurrence strength between classes. A Gated Graph Neural Network (GGNN) \cite{Li2016} is then applied to propagate information across this graph, enabling class-specific representations to be contextually refined based on their related classes.  





\noindent \textbf{Intra-image Semantic Transfer (IST).}
To refine label estimates within each sample, the IST module constructs a class co-occurrence matrix based on semantic-aware features. 
Although originally proposed for images, this module is employed for ECG signals without any architectural modifications. 
For a pair of classes $j$ and $k$, their co-occurrence probability is calculated as:
\begin{equation} 
p^{ist}_{j,k} = f^{ist}([\tilde{\phi}_{i,j}, \tilde{\phi}_{i,k}]),
\end{equation}
where $f^{ist}$ is a neural network composed of fully connected layers, and $[\cdot]$ denotes feature concatenation. This yields a matrix $P^{ist}_i \in \mathbb{R}^{C \times C}$ capturing the co-occurrence strength of class pairs. The refined label estimate for class $k$ is then given by:
\begin{equation}
\tilde{y}_k = \mathbf{1}\left[\left(\sum_{y_j \in Y_i} p^{ist}_{k,j} y_j\right) \geq \theta^{ist} \right],
\end{equation}
where $\theta^{ist}$ is a predefined threshold, and $\mathbf{1}[\cdot]$ is the indicator function. The IST module is trained using a loss function that encourages high co-occurrence scores for observed label pairs and penalizes unlikely prediction combinations:
\begin{equation}
L^{ist} = \sum_{(y_j, y_k) \in B_i} (1 - p^{ist}_{j,k})^{\gamma_1} \log(p^{ist}_{j,k}) + \sum_{(y_j, y_k) \notin B_i} (p^{ist}_{j,k} - m)^{\gamma_2} \log(1 - p^{ist}_{j,k}),
\end{equation}
where $B_i$ denotes the set of label pairs in the candidate set $Y_i$, and $\gamma_1$, $\gamma_2$, and $m$ are hyperparameters.

\noindent \textbf{Cross-image Semantic Transfer (CST).}
Complementing IST, the CST module captures inter-sample relationships, assuming that inputs with similar semantic content likely share labels. Similar to the IST module, although CST was originally proposed for images, it is employed for ECG signals without any architectural modifications.
For a given class $j$, the similarity between semantic-aware features of two inputs $x_i$ and $x_z$ is computed via cosine similarity:
\begin{equation}
s^j_{i,z} = \cos(\tilde{\phi}_{i,j}, \tilde{\phi}_{z,j}).
\end{equation}
The refined label for class $j$ in sample $x_i$ is then estimated as:
\begin{equation}
\widehat{y}_j^i = \mathbf{1}\left[\left( \frac{1}{|\mathcal{D}_j|} \sum_{x_z \in \mathcal{D}_j} s^j_{i,z} \right) \geq \theta^{cst} \right],
\end{equation}
where $\mathcal{D}_j$ is the set of samples containing class $j$, and $\theta^{cst}$ is a predefined similarity threshold. The CST loss encourages high similarity prediction between related ECG samples:
\begin{equation}
L^{cst} = \sum_{z=1}^{N} \sum_{y_j \in Y_i} (1 - s^j_{i,z}).
\end{equation}

\noindent \textbf{Classification Loss.}
SST employs a partial binary cross-entropy loss to handle ambiguous labels:
\begin{equation}
\ell(\mathbf{p}_i, Y_i) = \frac{1}{|Y_i|} \left( \sum_{y_j \in Y_i} \log(p_j^i) + \sum_{y_j \notin Y_i} \log(1 - p_j^i) \right).
\end{equation}
This loss is computed not only on the original candidate labels $Y_i$, but also on pseudo-labels obtained from the IST and CST modules, denoted as $\hat{Y}_i$ and $\tilde{Y}_i$, respectively. The final classification loss becomes:
\begin{equation}
L_{cls} = \sum_{i=1}^{N} \left( \ell(\mathbf{p}_i, Y_i) + \ell(\mathbf{p}_i, \hat{Y}_i) + \ell(\mathbf{p}_i, \tilde{Y}_i) \right).
\end{equation}
The total loss for the SST framework integrates all components:
\begin{equation}
\mathcal{L}_{SST} = L_{cls} + \lambda_1 L_{ist} + \lambda_2 L_{cst},
\end{equation}
where $\lambda_1$ and $\lambda_2$ control the contributions of the IST and CST modules, respectively. Accordingly, we apply the SST loss $\mathcal{L}_{SST}$ to train our ECG classification model under PLL setup.

\subsection{Heterogeneous Semantic Transfer }
Heterogeneous Semantic Transfer (HST) \cite{Chen2024a} builds upon the SST framework by introducing key enhancements aimed at improving efficiency and adaptability in PLL. While the backbone network and the IST module remain similar to those in SST, the CST module is significantly improved. Instead of computing pairwise similarities between samples, HST introduces class prototypes to efficiently capture inter-sample semantics.

For each class $j$, semantic-aware representations of samples containing class $j$ are clustered using k-means to form $k$ prototypes $\{\tilde{p}_{1,j}, \dots, \tilde{p}_{k,j}\}$. Given an input $x_i$, the similarity to class $j$ is computed by averaging cosine similarities between $\tilde{\phi}_{i,j}$ and the class prototypes:
\begin{equation}
s^{j,k}_{i} = \cos(\tilde{\phi}_{i,j}, \tilde{p}_{k,j}), \quad 
\widehat{y}_i^j = \mathbf{1}\left[\left(\frac{1}{K}\sum_{k=1}^{K} s^{j,k}_i\right) \geq \theta^{cst} \right].
\end{equation}
This strategy reduces computational cost while preserving semantic alignment. The CST loss remains similar to SST:
\begin{equation}
L^{cst} = \sum_{z=1}^{N} \sum_{y_j \in Y_i} (1 - s^j_{i,z}).
\end{equation}

\noindent \textbf{Differential Threshold Learning (DTL).}
To eliminate the need for manual threshold tuning in the IST and CST modules of SST model, HST introduces the DTL module. Traditional threshold-based pseudo-labeling is sensitive to the values of $ \theta^{ist} $ and $ \theta^{cst} $, which are often dataset-specific. DTL addresses this by reformulating thresholding as a differentiable process, enabling the thresholds to be learned end-to-end via gradient descent.

For IST and CST, DTL defines the threshold difference scores as:
\begin{equation}
    d^{j,IST}_i = \left(\sum_{y_j \in Y_i} p^{ist}_{k,j} y_j\right) - \theta^{ist}, \quad 
    d^{j,CST}_i = \left(\frac{1}{K} \sum_{k=1}^{K} s^{j,k}_i\right) - \theta^{cst}.
\end{equation}
These values are then used to supervise the threshold learning using partial binary cross-entropy:
\begin{equation}
    L_{ist}^{dtl} = \sum_{i=1}^{N} \ell\left(Y_i, \mathbf{d}_i^{IST}\right), \quad 
    L_{cst}^{dtl} = \sum_{i=1}^{N} \ell\left(Y_i, \mathbf{d}_i^{CST}\right),
\end{equation}
where $ \ell $ denotes the partial binary cross-entropy loss. By leveraging the candidate labels as supervision, DTL dynamically learns suitable thresholds that balance recall and precision, improving robustness to ambiguous labels.

\noindent \textbf{Classification Loss.}
Similar to SST, HST computes classification loss over the original candidate labels as well as the pseudo-labels generated by IST and CST. The total training objective for the HST model is:
\begin{equation}
\mathcal{L}_{HST} = L_{cls} + \lambda_1 L_{ist} + \lambda_2 L_{cst} + \lambda_3\left(L_{ist}^{dtl} + L_{cst}^{dtl}\right),
\end{equation}
where $ \lambda_1 $, $ \lambda_2 $, and $ \lambda_3 $ are hyperparameters that control the contribution of each component. Similar to SST, we apply the HST loss $\mathcal{L}_{HST}$ to train our ECG classification model, incorporating the enhanced CST design and adaptive threshold learning to improve efficiency and performance under partial label supervision.

\subsection{COrrection ModificatIon balanCe}
PLL methods often face significant challenges when dealing with datasets that exhibit a \textit{long-tailed} label distribution and multiple forms of imbalance. In such scenarios, two types of imbalance commonly occur: (1) a \textit{head–tail} (major–minor) class imbalance, where a few head classes have abundant samples while many tail classes are underrepresented; and (2) a \textit{positive–negative} imbalance, where each instance contains only a few positive labels but many negative labels. These imbalances, combined with label ambiguity, can cause models to overfit to frequent head classes, underfit rare tail classes, and be dominated by the overwhelming number of negative labels, leading to poor generalization. COrrection, ModificatIon, balanCe (COMIC) \cite{Zhang2023a} addresses this challenging setting by unifying three complementary components: (1) a class-frequency-aware loss function, named Multi-Focal Modifier (MFM), that incorporates both inter-instance head–tail balancing and intra-instance positive–negative balancing; (2) a parallel head–tail balanced learning architecture with cross-model knowledge distillation; and (3) a label correction mechanism that iteratively refines candidate label sets. The integration of these modules enables COMIC to simultaneously disambiguate candidate labels, mitigate both forms of imbalance, and enhance robustness under ambiguous supervision.

\noindent \textbf{MFM loss.}
To address imbalance in label frequency, COMIC adopts a modified Focal Loss \cite{Lin2017} tailored for long-tailed data. The MFM loss reduces emphasis on head classes while amplifying the gradient contribution from tail classes. This is achieved through dynamic scaling factors that combine intra-sample and inter-class imbalance metrics, ensuring that underrepresented classes receive proportionally higher attention. The intra-sample factor controls the positive-negative weighting within a sample, while the inter-class \textit{head–tail factor} increases with class imbalance degree, further emphasizing rare classes. Given predicted probability $p$ for the $j^{\text{th}}$ class, the MFM loss is defined separately for positive and negative labels as:
\begin{equation}
L_m^{+}=\sum_{j=1}^{C}(1-p_j)^{\gamma^+_{j}} \log (p_j), \quad
L_m^{-}=\sum_{j=1}^{C} p_j^{\gamma^-_{j}} \log (1-p_j),
\end{equation}
where $\gamma^+_{j}$ and $\gamma^-_{j}$ control sensitivity for positive and negative samples, respectively:
\begin{equation}
\gamma^+_{j}=\gamma^{pn+}+w^{+} \cdot \gamma^{ht}_{j}, \quad
\gamma^-_{j}=\gamma^{pn-}+w^{-} \cdot \gamma^{ht}_{j}.
\end{equation}
Here, $\gamma^{pn+}$ and $\gamma^{pn-}$ are intra-sample positive–negative scaling factors, with $\gamma^{pn-} \geq \gamma^{pn+}$ to upweight positive samples; $w^{+}$ and $w^{-}$ are coefficients that adjust inter-class weighting; and $\gamma^{ht}_{j} \geq 1$ is the head–tail imbalance factor reflecting the imbalance degree of class $j$ with higher values for minor classes. This loss formulation amplifies the gradient contribution of positive samples from tail classes while reducing overemphasis on head classes or negative samples.

\noindent \textbf{Head–Tail Balancer.}
Even with reweighted loss functions, extreme imbalance can cause head classes to overfit and tail classes to underfit. To counter this, COMIC trains three parallel models with different learning rate decays: a head-class expert $\mathcal{M}_h$, a tail-class expert $\mathcal{M}_t$, and a balanced model $\mathcal{M}_b$. While an ECG signal is passed through these models, the balanced model aggregates features from the two head and tail models via an additive attention mechanism, producing a shared representation for classification, as follows:
\begin{equation}
\phi_{b} = \text{Attn}\left(\hat{\phi}_{b},\,[\phi_{h}, \phi_{t}]\right) + \hat{\phi}_{b},
\end{equation}
where $\phi_{h}$, $\phi_{t}$, and $\hat{\phi}_{b}$ are the feature embeddings from $\mathcal{M}_h$, $\mathcal{M}_t$, and $\mathcal{M}_b$, respectively.

A multi-head classifier normalizes class logits by dividing the $k^{\text{th}}$ weight vector $w_k$ into $q$ groups:
\begin{equation}
\mathbf{z}_{x} = \frac{\rho}{q} \sum_{k=1}^{q} \frac{w_{k}^{\top} \phi_{x}}{(\|w_{k}\|+\eta)\|\phi_{x}\|}, \quad x \in\{h,t,b\},
\end{equation}
where $\rho$ is a scaling factor and $\eta$ is a class-agnostic baseline energy. 

To further enforce balance, the model’s optimization bias is estimated from the moving average of accumulated gradients:
\begin{equation}
\mathbf{e}_{t} = \mu \cdot \mathbf{e}_{t-1} + \text{sum}(g_{t}),
\end{equation}
where $\text{sum}(g_t)$ aggregates the gradients at iteration $t$. This vector $\mathbf{e}_t$ is added and subtracted respectively from the logits of $\mathcal{M}_h$ and $\mathcal{M}_t$ to simulate balanced training on major and minor classes:
\begin{equation}
\hat{\mathbf{z}}_{x} = \mathbf{z}_{x} \pm \frac{\rho}{q} \sum_{k=1}^{q} \frac{\text{sim}(\mathbf{z}_{x},\mathbf{e}_t)\cdot w_{j}^\top \mathbf{e}_t}{\|w_{k}\|+\eta}, \quad x\in\{h,t\},
\end{equation}
where sim measures the cosine similarity of the feature vectors.

Finally, knowledge distillation is applied from $\hat{\mathbf{z}}_h$ and $\hat{\mathbf{z}}_t$ to $\mathbf{z}_b$, with a balancing loss, encouraging uniform performance across all classes:
\begin{equation}
L_b = \kappa_{h} L_m\big(\sigma(\hat{\mathbf{z}}_{h}),\sigma(\mathbf{z}_{b})\big) + \kappa_{t} L_m\big(\sigma(\hat{\mathbf{z}}_{t}),\sigma(\mathbf{z}_{b})\big),
\end{equation}
where $\sigma$ is the softmax, and $\kappa_h,\kappa_t$ are adaptive weights:
\begin{equation}
\kappa_x=\frac{(\mathcal{L}(\hat{\mathbf{z}}_{x}))^{\alpha}}{(\mathcal{L}(\hat{\mathbf{z}}_{t}))^{\alpha}+(\mathcal{L}(\hat{\mathbf{z}}_{h}))^{\alpha}}, \quad x\in\{h,t\},
\end{equation}
where $\alpha$ is the scaling factor. This loss can be seen as an empirical risk minimization which distills knowledge from $\mathcal{M}_h$ and $\mathcal{M}_t$. To accommodate ECG inputs, we employ ResNet18 as the common feature extraction backbone for all three models, followed by a multi-head classifier as the final classification layer.

\noindent \textbf{Reflective Label Corrector.}
To improve label quality, COMIC iteratively refines the candidate label set based on model confidence. Labels with predicted probability exceeding a dynamic threshold, computed from both a global category prior and a fixed margin, are considered as positive labels, while others are either removed or retained with lower weight. This correction step is designed to be class-aware: it scales the loss contribution using the head–tail factor so that tail-class labels receive more influence during optimization. This prevents the correction process from disproportionately favoring head classes, which naturally tend to have higher confidence predictions.

Given candidate labels $Y_i$ for a sample, COMIC refines them using model confidence $p_c$. With threshold $\tau$ and per-class average probability $P_j$, the updated pseudo-label $\widehat{y}_j$ is:
\begin{equation}
\widehat{y}_{j}=\begin{cases}
1, & \text{if } p_{c}>\max \{\tau, P_{j}\},\ y_{j}=1,\\
0, & \text{otherwise}.
\end{cases}
\end{equation}
The classification loss for each class is then:
\begin{equation}
L_c(p_j) = \begin{cases}
L_m^{+}(p_j), & \widehat{y}_j=1,\\
\mathbb{1}_{(y_j=1)} L_m^{+}(p_j) + L_m^{-}(p_j), & \text{otherwise},
\end{cases}
\end{equation}
scaled by $\frac{B}{\mathcal{N}_t}$, where $B$ is batch size and $\mathcal{N}_t$ is the number of corrected labels. Since corrections tend to favor head classes, $L_c$ is additionally scaled by $\gamma^{ht}_i$ as the inter-sample head-tail factor to upweight tail classes.

\noindent \textbf{Total Loss.}
The total loss of the COMIC model combines the three core components, including MFM loss $L_m$, head–tail balancing loss $L_b$, and label correction loss $L_c$, as:
\begin{equation}
\mathcal{L}_{COMIC} = \lambda_m L_m + \lambda_b L_b+\lambda_c L_c  ,
\end{equation}
where $\lambda_m$, $\lambda_b$, and $\lambda_c$ control the contributions of modification, balancing, and correction respectively. This formulation enables COMIC to jointly perform noise-robust disambiguation and balanced learning in the presence of long-tailed, partially labeled datasets.

\section{Experiments}

\subsection{Dataset}
We conduct our experiments on three popular publicly available ECG datasets:

\noindent\textbf{(1) CODE Test.}
The CODE Test dataset \cite{Ribeiro2020} contains 827 12-lead ECG recordings annotated for six common cardiac conditions, including atrial fibrillation (AF), sinus bradycardia (SB), sinus tachycardia (ST), first-degree AV block (1dAVb), and bundle branch blocks (LBBB, RBBB). A key characteristic of this dataset is its multi-annotator labeling protocol. Each ECG is independently annotated by six clinicians with different levels of expertise, namely three cardiology experts, a cardiology resident, an emergency medicine resident, and a medical student. One of the cardiology experts serves as the gold standard for this dataset. By providing annotations from multiple clinicians with different levels of expertise, this dataset offers a unique opportunity to study diagnostic disagreement in complex ECG cases. 

\noindent\textbf{(2) PTB-XL.} The PTB-XL dataset \cite{Wagner2020} is a large-scale, publicly available clinical ECG corpus comprising 21,837 12-lead recordings from 18,885 patients. Each recording is annotated with one or more diagnostic labels, organized within a hierarchical taxonomy of five superclass categories and 24 subclass labels. In this study, we focus exclusively on the 24 subclass labels as a multi-label classification task.

\noindent\textbf{(3) Chapman (PhysioNet 2020 version).} The Chapman dataset \cite{Zheng2020} consists of 12-lead ECG recordings collected from 10,646 patients by Chapman University and Shaoxing People’s Hospital. Each recording spans 10 seconds and is sampled at 500 Hz, providing rich temporal information. For this study, we use the version released as part of the PhysioNet 2020 Challenge \cite{Alday2020}, which unifies diagnostic labels across multiple ECG datasets. 

\begin{figure*}
    \centering
    \begin{subfigure}{0.45\textwidth}
            \centering
            \includegraphics[width=0.9\linewidth]{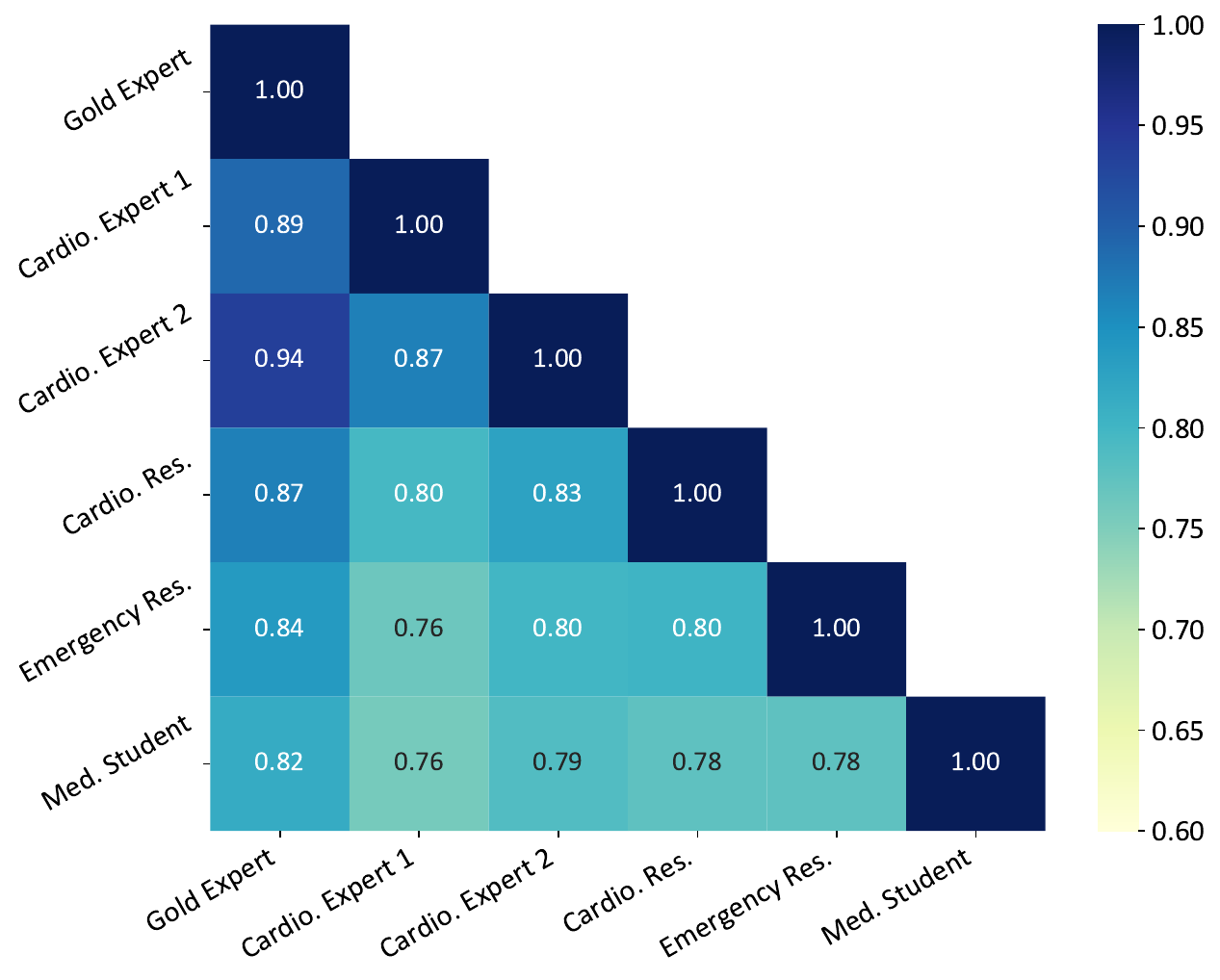}
            \caption{Pairwise Annotator Agreement (Cohen's Kappa)}
            \label{fig:annot_heat}
    \end{subfigure}
    \begin{subfigure}{0.48\textwidth}
            \centering
            \includegraphics[width=0.85\linewidth]{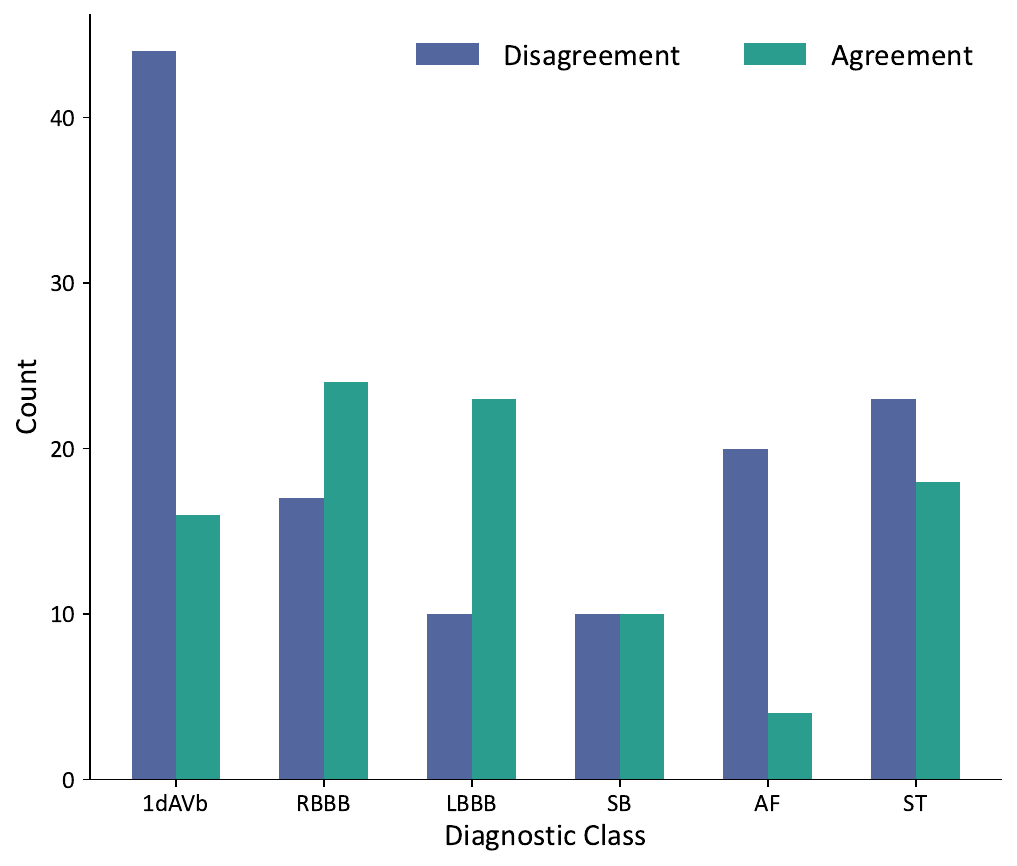}
            \caption{Class-wise Annotator Agreement and Disagreement}
            \label{fig:annot_bar}
    \end{subfigure}
    \caption{Disagreement analysis in the CODE Test dataset. Pairwise annotator agreement is summarized using Cohen's kappa, while class-wise agreements and disagreement counts show which diagnostic classes are more prone to diagnostic disagreement.}
    \label{fig:annot_fig}
\end{figure*}

\subsection{Candidate Labels}
In real-world ECG diagnosis, diagnostic ambiguity often arises because an ECG recording may exhibit overlapping conditions or because different practitioners interpret the same signal differently. In a PLL setup, each ECG recording is annotated not with its exact ground-truth label set, but with a \textit{candidate label set} that contains the true diagnosis in addition to one or more plausible labels. For example, one annotator may diagnose an ECG as AF, while another may assign ST. We define a candidate label set as $Y = \{AF, ST\}$ to preserve the ambiguity for training a PLL method. Unlike a classical supervised setup, which resolves disagreement through majority voting or by relying on a senior expert, PLL preserves the ambiguity by defining a candidate label set such as $Y = \{AF, ST\}$. In the following, we describe both the real-world clinical ambiguity setting and a number of synthetic ambiguity generation strategies used in our study.

\subsubsection{Real-World Clinical Ambiguity}
In this setup, we use the natural diagnostic disagreement in the CODE Test dataset \cite{Ribeiro2020} to model real-world ambiguity. Each ECG is annotated by six clinicians with different levels of expertise, including cardiologists, residents, and medical students. Following the standard practice in the CODE Test dataset, annotations by a senior cardiologist (gold-standard expert) are considered as the ground truth label set $\tilde{Y}_i$. For each ECG sample, we define the candidate label set $Y_i$ as the union of the labels assigned by the annotators, given that it always contains the ground-truth label set $\tilde{Y}_i$. An ECG is considered ambiguous when disagreement among annotators results in a candidate label set that contains plausible labels in addition to the ground truth provided by the gold-standard expert, i.e., $Y_i \neq \tilde{Y}_i$. We consider all combinations of annotations as possible candidate label sets.

\textit{Annotator Disagreement Study.} In the CODE Test dataset, 12.6\% of samples have diagnostic disagreements, with a mean Fleiss' kappa of 0.81. We visualize the details of agreements and disagreements in this dataset in Fig. \ref{fig:annot_fig}. The pairwise agreement heatmap presents Cohen's kappa between annotators, while the class-wise disagreement plot shows how often annotators agree or disagree for each diagnostic class. Overall, the figure shows that agreement varies across annotator pairs and that disagreement is not uniformly distributed across diagnostic classes, where some classes show higher disagreement than others. This analysis highlights the value of CODE Test as a real-world ambiguity setting, where candidate labels reflect both annotator-level variability and class-specific diagnostic uncertainty.

\subsubsection{Synthetic Ambiguity}
\label{sec:ambiguityGeneration}
To construct candidate label sets in a controlled manner, we follow the PLL dataset generation protocol \cite{Tian2023}, where three parameters control introduction of label ambiguity: (1) $p \in [0,0.8]$ which is the proportion of samples converted into partial label form, (2) $r \in \mathbb{N}$, an optional fixed number that determines the number of incorrect labels added to the candidate set, and (3) $\epsilon \in [0,1]$, the probability of including any given incorrect label in the candidate set. Given a ground truth label set $\tilde{Y}_i$, a candidate label set $Y_i$ is generated as:
\begin{equation}
Y_i = \tilde{Y}_i \cup \left\{ y_j \in \mathcal{Y} \setminus \tilde{Y}_i \;\middle|\; \mathbb{I}_i \cdot \mathrm{Bernoulli}(\epsilon_{i,j}) = 1 \right\},
\end{equation}
where $\epsilon_{i,j}$ is the probability of including an incorrect label $y_j$, and $\mathbb{I}_i \sim \mathrm{Bernoulli}(p)$ is an indicator denoting whether $x_i$ is annotated with partial labels. The probability $\epsilon_{i,j}$ can be defined uniformly across classes (symmetric ambiguity, for instance randomly assigned partial labels) or conditioned on \textit{Class-} or \textit{Instance-Level} properties (structured ambiguity). 

This formulation provides a principled mechanism for simulating diverse sources of diagnostic uncertainty, ranging from uninformative annotation errors to clinically meaningful confusions. In the following, we describe the various ambiguity generation strategies.

\textbf{Random Ambiguity. }In this setting, incorrect labels are introduced uniformly across all classes, independent of semantic or structural relations. For each negative label $y_j$, the inclusion probability is fixed as $\epsilon_{i,j} = \epsilon$, where $\epsilon$ is a user-defined ambiguity rate. This mechanism mimics uninformative annotation noise and serves as widely adopted benchmark in PLL literature.

\begin{figure}
    \centering
    \includegraphics[width=0.9\linewidth]{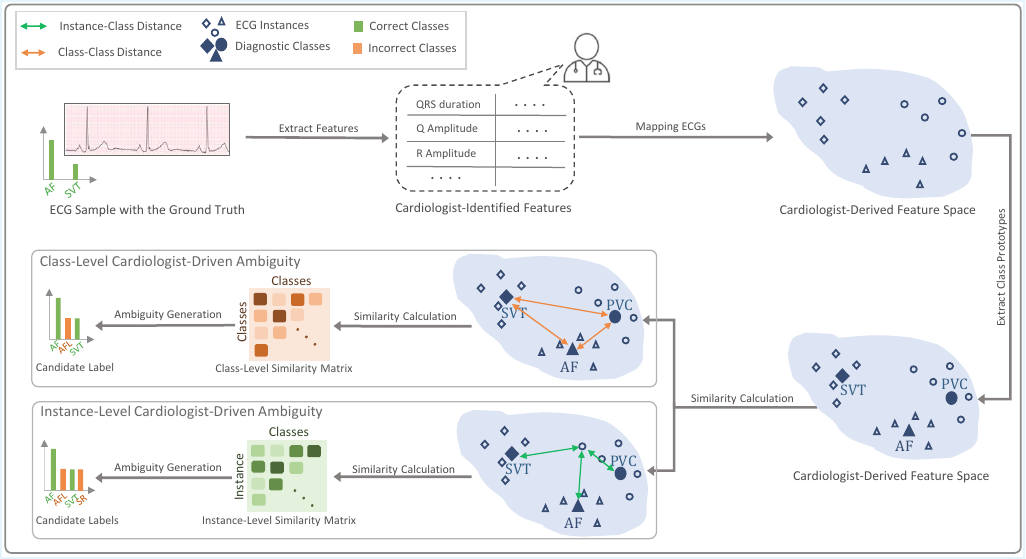}
    \caption{Cardiologist-derived ambiguity modeling for ECG datasets. Each ECG recording is represented by clinically significant diagnostic features such as QRS duration, Q amplitude, and R amplitude. These features form a cardiologist-derived feature space, where class prototypes represent each diagnostic class. Pairwise cosine similarities between class prototypes create a \textit{Class-Level} similarity matrix (orange), reflecting global confusion between diagnostic classes. Similarly, the cosine similarities between each individual ECG and class prototypes produce an \textit{Instance-Level} similarity matrix (green), which captures patient-specific diagnostic ambiguity.}
    \label{fig:cardiopercept}
\end{figure}

\textbf{Class-Level Ambiguity. }To model ambiguity arising from systematic inter-class confusions in real-world scenarios, we create \textit{Class-Level} ambiguity guided by similarities between classes of disease. Let $T \in \mathbb{R}^{C \times C}$ denote a \textit{Class-Level} transition matrix, where $t_{k,z}$ represents the probability of mislabeling class $k$ as class $z$. For an instance $x_i$ with ground-truth labels $\tilde{Y}_i$, the probability of including label $z$ in the candidate label set, is defined as
\begin{equation}
\epsilon_{i,z} = \frac{1}{|\tilde{Y}_i|} \sum_{y_k \in \tilde{Y}_i} t_{k,z}.
\end{equation}
This formulation introduces ambiguity proportionally to inter-class similarity. We instantiate $T$ under the following variants:
\begin{itemize}
    \item \textbf{Treatment-Driven Ambiguity:} Classes with overlapping treatment protocols are likely to co-occur or be confused in practice. We adopt the treatment similarity matrix from the PhysioNet 2020 Challenge \cite{Alday2020} and use it directly as $T$.

    \item \textbf{Class-Level Cardiologist-Driven Ambiguity:} To incorporate domain expertise into ambiguity generation, we construct \textit{Class-Level} similarity matrices using features that cardiologists have identified as diagnostically significant for disease classification \cite{Mehari2024}.    
    Each ECG sample $x_i$ is mapped to a feature vector $\psi(x_i)$ containing clinically significant diagnostic attributes identified in \cite{Mehari2024}. 
    For each diagnosis class $y_j \in \mathcal{Y}$, we then compute a class prototype by averaging the feature vectors of all ECGs belonging to that class:
    \begin{equation}
        v_j = \frac{1}{|\mathcal{D}_j|} \sum_{x_i \in \mathcal{D}_j} \psi(x_i),
    \end{equation}
    where $\mathcal{D}_j$ is the set of all samples with ground-truth label $y_j$. Pairwise similarities between class prototypes are computed using cosine similarity, resulting in the transition matrix $T$ that captures clinically meaningful inter-class relationships (illustrated in Fig. \ref{fig:cardiopercept}).
    Since raw cosine similarities vary in scale, we apply min–max normalization to ensure comparability across classes:
    \begin{equation}
    t_{j,k} = \frac{\hat{t}_{j,k} - \min(T)}{\max(T) - \min(T)}.
    \end{equation}
    
    This normalization maps similarity values into a standardized range of $[0,1]$ while preserving relative differences between classes. The resulting matrix $T$ highlights cardiologist-derived diagnostic ambiguities, which is subsequently used to guide candidate label generation.

    \item \textbf{Taxonomy-Driven Ambiguity:}  
    To incorporate domain taxonomies into ambiguity generation, we restrict candidate label ambiguity to sibling classes within the diagnostic hierarchy. Intuitively, ECG classes that share the same superclass (e.g., IRBBB and CRBBB) are more likely to be misdiagnosed in practice.

    Formally, for each class $y_j \in \mathcal{Y}$, the transition probabilities in $T$ are defined as:
    \begin{equation}
    t_{j,k} = 
    \begin{cases}
    \epsilon, & \text{if } \;\text{Sup}(y_j) = \text{Sup}(y_k), \\
    0, & \text{otherwise},
    \end{cases}
    \end{equation}
    where $\text{Sup}(y_j)$ denotes the superclass of $y_j$ in the diagnostic taxonomy, and $\epsilon$ is a fixed ambiguity rate (set to $0.5$ in our experiments). This formulation ensures that ambiguity reflects clinically plausible confusion patterns while avoiding unrealistic cross-superclass errors.

\end{itemize}

\textbf{Instance-Level Ambiguity. }To account for ambiguity at the individual patient level, we further extend label ambiguity modeling to \textit{Instance-Level} distributions. Let $T \in \mathbb{R}^{n \times C}$ denote an \textit{Instance-Level} transition matrix, where $t_{i,k}$ is the probability of including class $k$ for instance $x_i$. We consider two biologically motivated variants of this type of ambiguity:

\begin{itemize}

    \item \textbf{Instance-Level Cardiologist-Driven Ambiguity:}  
    To capture ambiguity at the level of individual ECG recordings, we extend the \textit{Class-Level} Cardiologist-Derived similarity to an \textit{Instance-Level} formulation. For each ECG $x_i$, we extract its feature vector $\psi(x_i)$ based on cardiologist-identified diagnostic attributes \cite{Mehari2024}. This instance representation is then compared with each class prototype $v_j$ (Fig. \ref{fig:cardiopercept}) using cosine similarity, producing a similarity distribution $t_{i,:}$ over all classes. Finally, the scores are standardized via the min–max normalization procedure. 

    \item \textbf{Model–Driven Ambiguity:} Inspired by pseudo-labeling techniques in PLL \cite{Xu2021,Xu2024}, we leverage the confidence scores of a classifier $f(\cdot;\theta)$ trained on clean ground-truth labels to simulate ambiguity. For a given instance $x_i$ and an incorrect class $y_k$, the \textit{Instance-Level} transition matrix is defined as
    \begin{equation}
        t_{i,k} =         
        \frac{f_k(x_i;\theta)}{\max_{y_j \in \mathcal{Y}\setminus\tilde{Y}_i} f_j(x_i;\theta)}\;\;\;\;\;\;\; \text{for}\;\;y_k \in \mathcal{Y}\setminus\tilde{Y}_i,
    \end{equation}
    where $f_j(x_i;\theta)$ denotes the predicted probability for class $y_j$. This formulation scales the confidence of each incorrect class relative to the most confident incorrect prediction, yielding a normalized distribution over potential false positives.

\end{itemize}

\subsection{Implementation and Evaluation Details}
All baseline models are implemented within a unified framework to ensure fair and consistent comparison. We adopt a 1D ResNet18 as the shared backbone architecture for all PLL methods, composed of four stages with two bottleneck blocks each. Each bottleneck block contains three convolutional layers: a 1×1 convolution for channel reduction, a 3×3 convolution (stride 1 for the first stage and 2 for the remaining stages) with padding 1, and a final 1×1 convolution to restore channels. Batch normalization and ReLU activation follow each convolutional layer, except the last, where ReLU is applied after residual addition. The final feature map is reduced via adaptive average pooling to a single value per channel and passed through a fully connected layer to produce the outputs. Training is conducted using a batch size of 32 for PTB-XL and Chapman, and 22 for CODE Test, aligned with the dataset's split size to maximize sample utilization per epoch, a learning rate of 0.001, and the RMSprop optimizer with default momentum settings over 20 epochs. Each method follows its original network and parameter configuration as closely as possible. All experiments are conducted in PyTorch on an NVIDIA RTX 2080 GPU. 

To evaluate performance under partial supervision, we adopt micro-F1 and AUROC as widely used metrics in ECG classification \cite{Zhang2025a}. For all datasets, we use an 80/20 train-test split, with 80\% of samples used for training and 20\% held out for testing. The results for synthetic ambiguity generation on PTB-XL and Chapman dataset are reported on a clean held-out test set, while training is performed on versions of the training set augmented with varying levels of label ambiguity introduced by the six candidate label generation strategies described in Section~\ref{sec:ambiguityGeneration}. We repeat each experiment across three random seeds, and report the mean and standard deviation of both metrics to assess robustness. The full code will be made publicly available upon publication.

For the PTB-XL dataset, we apply all candidate label generation strategies described in Section~\ref{sec:ambiguityGeneration}. For the \textit{Class-Level} \textit{Treatment-Driven} strategy, we adopt label definitions from the PhysioNet 2020 Challenge \cite{Alday2020} to leverage clinically curated inter-class dependency weights, enabling a more realistic modeling of diagnostic ambiguity. For the \textit{Cardiologist-Driven} strategies, we use the diagnostic features provided for PTB-XL in \cite{Strodthoff2023}. In contrast, for the Chapman dataset (PhysioNet 2020 version), the available label schema supports only a subset of ambiguity strategies. Specifically, we generate candidate labels using the \textit{Treatment-Driven}, \textit{Random}, and \textit{Model-Driven} ambiguity strategies, while other variants that rely on explicit hierarchical taxonomies or cardiologist crafted features are not applicable to this dataset.

\begin{table*}[t]
    \centering
    \caption{Comparison of PLL baseline models under natural real-world clinical ambiguity on CODE Test, including overall micro-F1 and class-specific AUROC performance. Bold numbers denote the best performance, while underline indicates the second best.}
    \renewcommand{\arraystretch}{1.2}
    \begin{adjustbox}{max width=\textwidth}
    \begin{tabular}{l|c|ccccccc}
        \toprule
        \multirow{2}{*}{\textbf{Baseline Models}}
        & \multicolumn{1}{c|}{\textbf{Overall}}
        & \multicolumn{7}{c}{\textbf{Class-Specific Performance}} \\
        & \textbf{Micro-F1}
        & \textbf{1dAVb}
        & \textbf{RBBB}
        & \textbf{LBBB}
        & \textbf{SB}
        & \textbf{AF}
        & \textbf{ST}
        & \textbf{NORM} \\
        \midrule
        \midrule
        \textbf{No PLL} & 0.709{\footnotesize ±0.129} & 0.591{\footnotesize ±0.133} & 0.743{\footnotesize ±0.145} & 0.741{\footnotesize ±0.118} & 0.500{\footnotesize ±0.000} & 0.500{\footnotesize ±0.000} & 0.705{\footnotesize ±0.156} & 0.599{\footnotesize ±0.023} \\
        
        \midrule
        \textbf{DNPL \cite{Seo2021}} & \textbf{0.779{\footnotesize ±0.003}} & 0.513{\footnotesize ±0.052} & 0.782{\footnotesize ±0.008} & 0.772{\footnotesize ±0.002} & 0.600{\footnotesize ±0.073} & 0.500{\footnotesize ±0.000} & 0.633{\footnotesize ±0.024} & 0.542{\footnotesize ±0.011} \\
        
        \textbf{PRODEN \cite{Lv2020}} & 0.756{\footnotesize ±0.014} & 0.538{\footnotesize ±0.040} & 0.804{\footnotesize ±0.035} & 0.815{\footnotesize ±0.033} & \textbf{0.641{\footnotesize ±0.007}} & 0.498{\footnotesize ±0.003} & \textbf{0.850{\footnotesize ±0.040}} & 0.537{\footnotesize ±0.013} \\
        
        \textbf{CAVL \cite{Zhang2021}} & 0.710{\footnotesize ±0.026} & \textbf{0.616{\footnotesize ±0.029}} & 0.740{\footnotesize ±0.100} & 0.498{\footnotesize ±0.003} & 0.500{\footnotesize ±0.000} & 0.500{\footnotesize ±0.000} & 0.549{\footnotesize ±0.041} & 0.523{\footnotesize ±0.007} \\
        
        \textbf{LW \cite{Wen2021}} & 0.766{\footnotesize ±0.009} & 0.504{\footnotesize ±0.030} & 0.750{\footnotesize ±0.120} & 0.734{\footnotesize ±0.072} & 0.550{\footnotesize ±0.071} & 0.500{\footnotesize ±0.000} & 0.798{\footnotesize ±0.073} & 0.534{\footnotesize ±0.029} \\
        
        \textbf{CR \cite{Wu2022}} & 0.593{\footnotesize ±0.061} & 0.529{\footnotesize ±0.088} & 0.684{\footnotesize ±0.088} & 0.668{\footnotesize ±0.058} & 0.467{\footnotesize ±0.063} & 0.500{\footnotesize ±0.000} & 0.685{\footnotesize ±0.037} & 0.519{\footnotesize ±0.017} \\
        
        \textbf{PICO \cite{Wang2022a}} & 0.692{\footnotesize ±0.012} & 0.521{\footnotesize ±0.047} & 0.606{\footnotesize ±0.042} & 0.780{\footnotesize ±0.025} & 0.462{\footnotesize ±0.015} & 0.500{\footnotesize ±0.000} & 0.719{\footnotesize ±0.071} & 0.502{\footnotesize ±0.017} \\
        
        \textbf{SST \cite{Chen2022a}} & \underline{0.771{\footnotesize ±0.003}} & 0.500{\footnotesize ±0.000} & \textbf{0.966{\footnotesize ±0.010}} & 0.642{\footnotesize ±0.053} & 0.500{\footnotesize ±0.000} & 0.499{\footnotesize ±0.002} & 0.499{\footnotesize ±0.002} & 0.659{\footnotesize ±0.010} \\
        
        \textbf{HST \cite{Chen2024a}} & 0.771{\footnotesize ±0.002} & 0.500{\footnotesize ±0.000} & 0.883{\footnotesize ±0.042} & 0.748{\footnotesize ±0.052} & 0.500{\footnotesize ±0.000} & 0.500{\footnotesize ±0.000} & 0.633{\footnotesize ±0.188} & \textbf{0.660{\footnotesize ±0.004}} \\
        
        \textbf{COMIC \cite{Zhang2023a}} & 0.523{\footnotesize ±0.022} & 0.580{\footnotesize ±0.007} & 0.774{\footnotesize ±0.065} & \textbf{0.875{\footnotesize ±0.082}} & 0.551{\footnotesize ±0.079} & \textbf{0.586{\footnotesize ±0.054}} & 0.763{\footnotesize ±0.060} & 0.532{\footnotesize ±0.017} \\
        \bottomrule
    \end{tabular}
    \end{adjustbox}
    \label{tab:code_realworld_results}
\end{table*}

\subsection{Performance Comparison}

Tables \ref{tab:synthetic_results} and \ref{tab:code_realworld_results} report the classification performance (Micro-F1 score) of nine state-of-the-art PLL models, along with a \textit{No PLL} baseline, under synthetic and real-world ambiguity settings. Table~\ref{tab:synthetic_results} summarizes the controlled settings on PTB-XL and Chapman, including \textit{Random}, \textit{Instance-Level} (\textit{Cardiologist-Driven} and \textit{Model-Driven}), and \textit{Class-Level} (\textit{Cardiologist-Driven}, \textit{Taxonomy-Driven}, and \textit{Treatment-Driven}) ambiguity. Table~\ref{tab:code_realworld_results} reports the natural real-world clinical ambiguity results on CODE Test, including both overall micro $F_1$ and class-specific AUROC performance. The controlled experiments are conducted on the PTB-XL and Chapman ECG datasets with 50\% of the training samples partially labeled ($p=0.5$). This comprehensive evaluation provides insight into the robustness of each method under different ambiguity structures.

Table \ref{tab:code_realworld_results} presents the \textit{Clinical Diagnosis Ambiguity} setting derived from the CODE Test dataset. In the real-world CODE Test setting, DNPL achieves the best overall performance, followed closely by SST and LW. Importantly, DNPL improves substantially over the \textit{No PLL} baseline, with a 0.07 gain in micro-F1. This is clinically meaningful because it suggests that methods relying directly on ambiguous supervision, including candidate and non-candidate label information, can outperform standard supervised training without needing complex disambiguation mechanisms on diagnostic disagreements. At the same time, the performance variation across SST, HST, and COMIC may indicate that real-world diagnostic disagreement does not necessarily follow a consistent ambiguity structure, which can challenge methods that rely on structural assumptions such as class relationships. The class-specific results further show that the strongest method varies across diagnostic classes, reflecting different scales and characteristics of ambiguity present in real-world clinical annotations (Fig. \ref{fig:annot_fig}). By comparing the class-specific performance to the annotator agreement and disagreement counts per class (Fig. \ref{fig:annot_bar}), we observe that all baselines perform better on classes with higher levels of agreement, such as RBBB, LBBB, and ST, where PRODEN shows consistent improvement in performance. Furthermore, PLL baselines provide only marginal improvement over classes where disagreement largely outweighs agreement, such as AF and 1dAVb, showing the difficulty of learning from highly ambiguous candidate label sets. Interestingly, COMIC performs stronger on AF compared to its weaker overall results, highlighting the effectiveness of its correction and class-balancing components under high ambiguity and limited training samples.

\begin{table*}[t]
    \centering
    \caption{Comparison of PLL baseline models under controlled synthetic ambiguity settings on PTB-XL and Chapman based on overall micro-F1. Bold numbers denote the best performance, while underline indicates the second best.}
    \renewcommand{\arraystretch}{1}
    \begin{adjustbox}{max width=\textwidth}
    \begin{tabular}{l|c|cc|ccc|ccc}
        \toprule
        \multirow{3}{*}{\textbf{Baselines}}
        & \multicolumn{6}{c|}{\textbf{PTB-XL}}
        & \multicolumn{3}{c}{\textbf{Chapman}} \\
        & \textbf{Random}
        & \multicolumn{2}{c|}{\textbf{Instance-Level}}
        & \multicolumn{3}{c|}{\textbf{Class-Level}}
        & \textbf{Random}
        & \textbf{Instance-Level}
        & \textbf{Class-Level} \\
        & -
        & {\small Card.-Driven}
        & {\small Model-Driven}
        & {\small Card.-Driven}
        & {\small Tax.-Driven}
        & {\small Treat.-Driven}
        & -
        & {\small Model-Driven}
        & {\small Treat.-Driven} \\
        \midrule
        \midrule
        \textbf{No PLL} & 0.432{\footnotesize ±0.024} & 0.383{\footnotesize ±0.042} & 0.289{\footnotesize ±0.012} & 0.390{\footnotesize ±0.044} & 0.611{\footnotesize ±0.008} & 0.520{\footnotesize ±0.041} & 0.543{\footnotesize ±0.045} & 0.395{\footnotesize ±0.024} & 0.556{\footnotesize ±0.046} \\
        \midrule
        \textbf{DNPL \cite{Seo2021}} & \underline{0.621{\footnotesize ±0.001}} & \underline{0.609{\footnotesize ±0.004}} & \underline{0.617{\footnotesize ±0.003}} & \underline{0.620{\footnotesize ±0.004}} & 0.615{\footnotesize ±0.006} & 0.677{\footnotesize ±0.010} & \underline{0.726{\footnotesize ±0.009}} & \underline{0.728{\footnotesize ±0.012}} & 0.730{\footnotesize ±0.006} \\
        \textbf{PRODEN \cite{Lv2020}} & \textbf{0.638{\footnotesize ±0.008}} & \textbf{0.637{\footnotesize ±0.008}} & \textbf{0.624{\footnotesize ±0.009}} & \textbf{0.637{\footnotesize ±0.011}} & 0.638{\footnotesize ±0.002} & \underline{0.747{\footnotesize ±0.001}} & 0.671{\footnotesize ±0.010} & 0.689{\footnotesize ±0.008} & 0.689{\footnotesize ±0.007} \\
        \textbf{CAVL \cite{Zhang2021}} & 0.595{\footnotesize ±0.003} & 0.586{\footnotesize ±0.006} & 0.594{\footnotesize ±0.006} & 0.583{\footnotesize ±0.011} & 0.612{\footnotesize ±0.003} & 0.685{\footnotesize ±0.004} & 0.656{\footnotesize ±0.019} & 0.712{\footnotesize ±0.014} & 0.684{\footnotesize ±0.010} \\
        \textbf{LW \cite{Wen2021}} & 0.599{\footnotesize ±0.008} & 0.593{\footnotesize ±0.006} & 0.608{\footnotesize ±0.004} & 0.599{\footnotesize ±0.003} & 0.590{\footnotesize ±0.008} & 0.736{\footnotesize ±0.001} & 0.725{\footnotesize ±0.012} & 0.719{\footnotesize ±0.011} & 0.708{\footnotesize ±0.001} \\
        \textbf{CR \cite{Wu2022}} & 0.534{\footnotesize ±0.013} & 0.506{\footnotesize ±0.016} & 0.526{\footnotesize ±0.020} & 0.524{\footnotesize ±0.013} & 0.530{\footnotesize ±0.008} & 0.730{\footnotesize ±0.003} & 0.691{\footnotesize ±0.014} & 0.682{\footnotesize ±0.013} & 0.701{\footnotesize ±0.017} \\
        \textbf{PICO \cite{Wang2022a}} & 0.591{\footnotesize ±0.008} & 0.594{\footnotesize ±0.002} & 0.541{\footnotesize ±0.013} & 0.585{\footnotesize ±0.003} & 0.554{\footnotesize ±0.003} & 0.690{\footnotesize ±0.004} & \textbf{0.740{\footnotesize ±0.006}} & \textbf{0.734{\footnotesize ±0.006}} & 0.737{\footnotesize ±0.002} \\
        \textbf{SST \cite{Chen2022a}} & 0.474{\footnotesize ±0.047} & 0.409{\footnotesize ±0.039} & 0.401{\footnotesize ±0.025} & 0.502{\footnotesize ±0.043} & 0.632{\footnotesize ±0.009} & 0.694{\footnotesize ±0.024} & 0.614{\footnotesize ±0.066} & 0.510{\footnotesize ±0.081} & \underline{0.761{\footnotesize ±0.010}} \\
        \textbf{HST \cite{Chen2024a}} & 0.516{\footnotesize ±0.015} & 0.498{\footnotesize ±0.048} & 0.293{\footnotesize ±0.025} & 0.545{\footnotesize ±0.011} & \underline{0.644{\footnotesize ±0.010}} & 0.717{\footnotesize ±0.023} & 0.718{\footnotesize ±0.053} & 0.347{\footnotesize ±0.027} & \textbf{0.773{\footnotesize ±0.005}} \\
        \textbf{COMIC \cite{Zhang2023a}} & 0.532{\footnotesize ±0.020} & 0.542{\footnotesize ±0.016} & 0.320{\footnotesize ±0.017} & 0.525{\footnotesize ±0.026} & \textbf{0.687{\footnotesize ±0.005}} & \textbf{0.754{\footnotesize ±0.003}} & 0.643{\footnotesize ±0.036} & 0.371{\footnotesize ±0.007} & 0.751{\footnotesize ±0.033} \\
        \bottomrule
    \end{tabular}
    \end{adjustbox}
    \label{tab:synthetic_results}
\end{table*}


On the Chapman dataset (Table \ref{tab:synthetic_results}), we observe a different trend. PICO consistently ranks among the best-performing methods, while simple approaches such as DNPL and LW also remain competitive. Notably, all PLL methods outperform the \textit{No PLL} baseline, validating the usefulness of considering label ambiguities. The strongest gains are observed under \textit{Treatment-Driven} ambiguity, where several methods (including COMIC, HST, and SST) outperform their results under \textit{Random} or \textit{Instance-Level} ambiguity. This suggests that when candidate labels reflect clinically grounded confusion patterns, PLL models can more effectively leverage structure to disambiguate. However, the instability of HST and SST persists under the more difficult ambiguity settings.

On PTB-XL (Table \ref{tab:synthetic_results}), most PLL methods outperform the \textit{No PLL} baseline, confirming the benefit of equipping models with PLL components under partial supervision. Among them, PRODEN delivers the most consistent and competitive performance across nearly all ambiguity scenarios. Its robustness highlights the effectiveness of iterative label distribution refinement for managing ambiguity in multi-label ECG diagnosis. LW and CR exhibit relatively strong performance, especially in scenarios involving \textit{Class-Level} ambiguities, suggesting that their reliance on negative supervision and consistency regularization effectively enhances label disambiguation. By contrast, SST and HST display notable degradation under \textit{Instance-Level} ambiguity, underscoring their limited adaptability beyond image-based assumptions. This could be attributed to their dependence on semantic embeddings for class relation extraction, which constrains their effectiveness when the embeddings are less informative. COMIC presents an interesting contrast: although weaker under random and \textit{Instance-Level} ambiguities, it achieves top performance under Treatment-Driven ambiguity. This demonstrates the benefit of its correction–modification mechanism when ambiguities simulate clinical relationships.

Overall, we make the following observations. First, while certain methods perform better in specific ambiguity scenarios, models that maintain stable performance across multiple conditions (such as DNPL or PRODEN) are more desirable for real-world ECG applications with varying types of ambiguity. Second, the differing trends observed across datasets suggest that dataset characteristics (e.g., class granularity, label distribution, or inherent noise) can significantly impact the effectiveness of PLL methods. This highlights the importance of benchmarking PLL frameworks across diverse datasets and ambiguity structures to fully understand their robustness. Most importantly, the CODE Test results show that methods that appear competitive under controlled synthetic ambiguity may still struggle when faced with clinically grounded annotation disagreement. In particular, methods such as COMIC and CR perform substantially worse on CODE Test than their synthetic results alone would suggest, indicating that annotation disagreement contains patterns that are not fully captured by controlled ambiguity generation. This clinically important finding suggests that real annotation disagreement reflects richer and less regular diagnostic uncertainty arising from expert variation in practice. Finally, the results of the \textit{No PLL} baseline highlight the negative impact of not considering label ambiguities. This emphasizes the necessity of integrating PLL frameworks, or similar ambiguity-aware learning strategies, into practical diagnostic systems.

\begin{figure}
    \centering
    \includegraphics[width=0.9\linewidth]{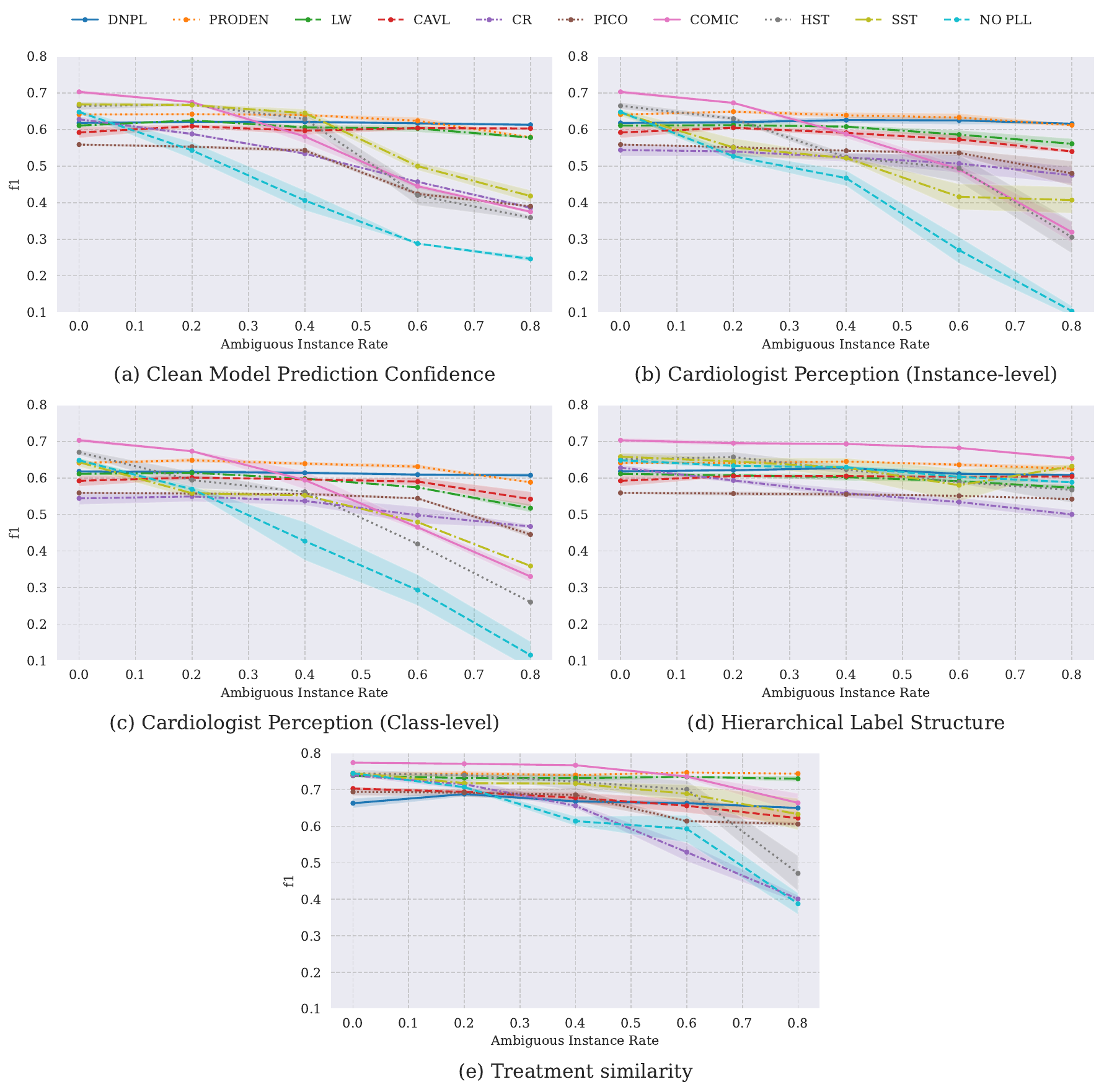}
    \caption{Baseline performance over different proportions of ambiguous samples ($p$) for PTB-XL dataset.}
    \label{fig:PTBBaselineperf}
\end{figure}
\begin{figure}
    \centering
    \includegraphics[width=0.9\linewidth]{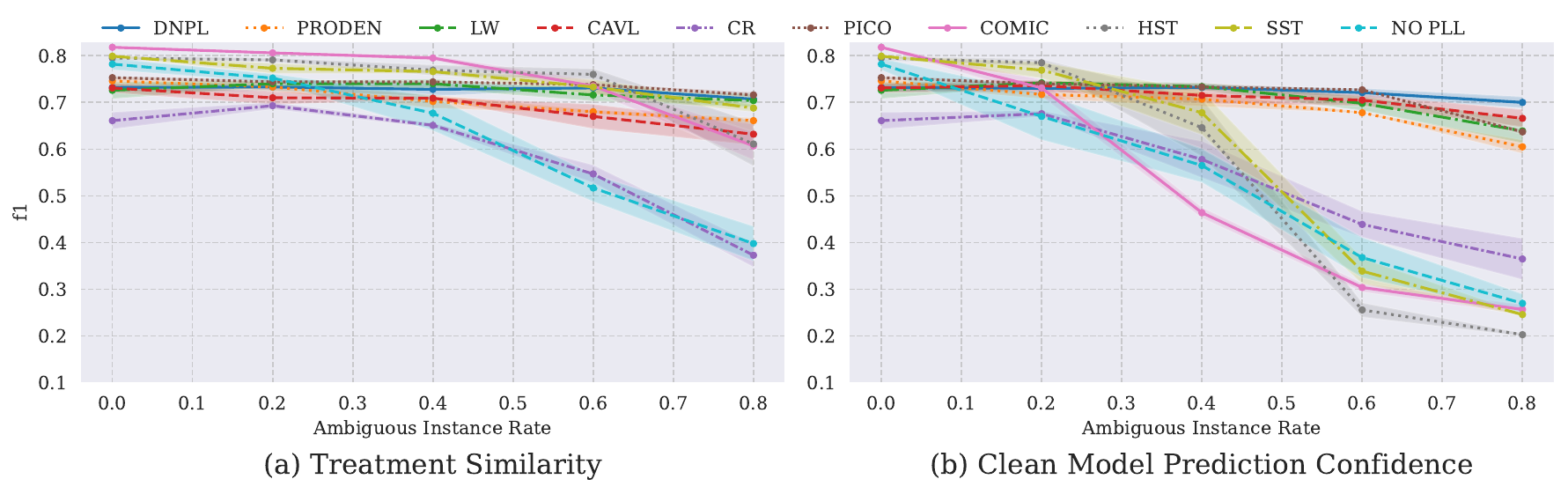}
    \caption{Baseline performance over different proportions of ambiguous samples ($p$) for Chapman dataset.}
    \label{fig:ChapmanBaselineperf}
\end{figure}

\begin{figure}
    \centering
    \includegraphics[width=0.9\linewidth]{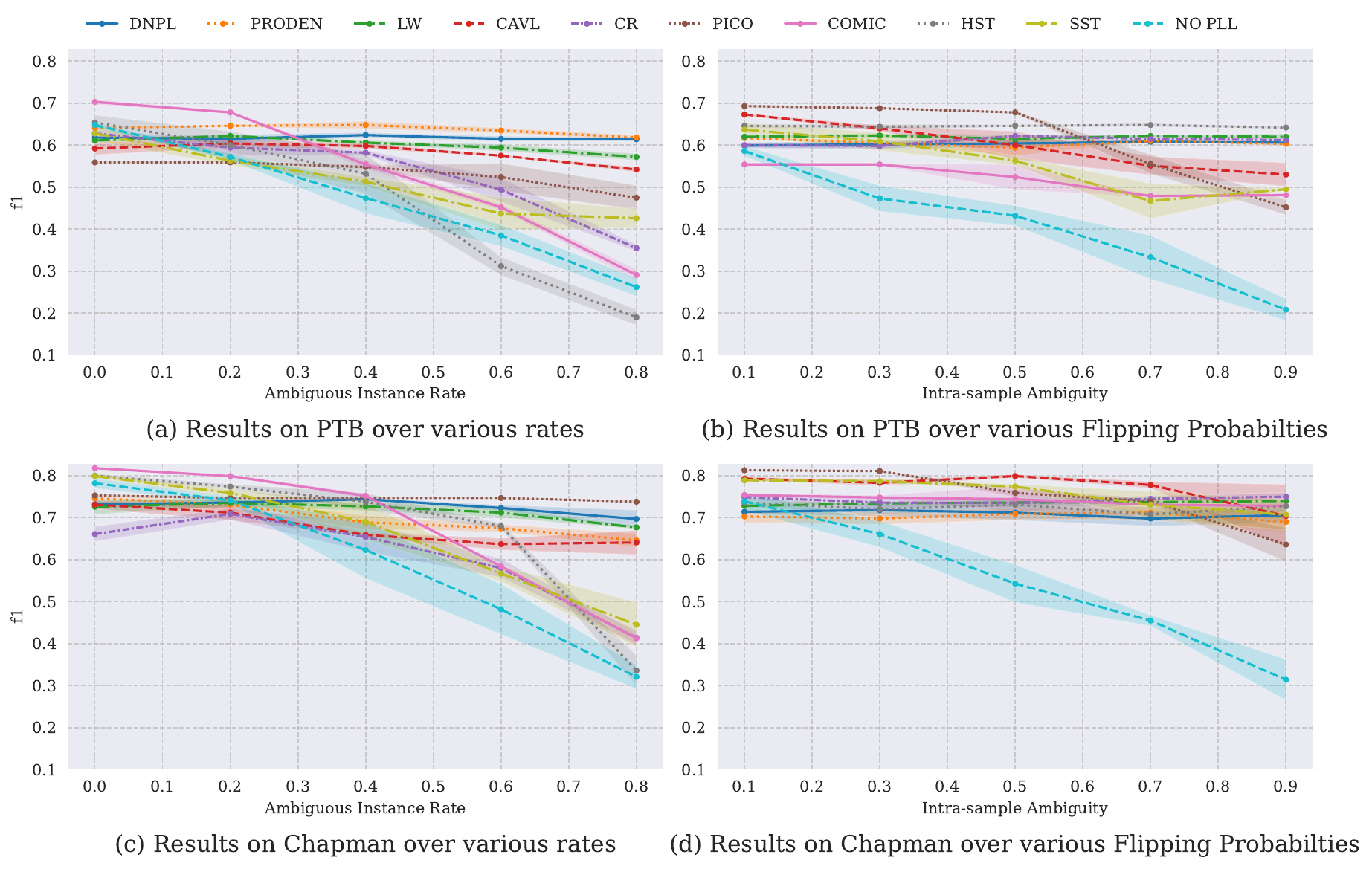}
    \caption{Baseline performance on random ambiguity over different proportions of ambiguous samples ($p$) and rates of ambiguity.}
    \label{fig:random_noise}
\end{figure}

\subsection{Sensitivity Analysis to Partial Label Proportion}
\label{sec:sensitivity_analysis}

To further investigate the robustness of baseline models to varying levels of label ambiguity, we conduct a series of controlled experiments by varying the proportion of ambiguous instances ($p \in [0,0.8]$). The corresponding results for the PTB-XL and Chapman datasets are presented in Figures~\ref{fig:PTBBaselineperf} and~\ref{fig:ChapmanBaselineperf}, respectively, while Figure~\ref{fig:random_noise} provides an additional analysis with \textit{Random} ambiguity by jointly varying the false-label inclusion probability ($\epsilon$) and the ambiguity proportion ($p$).
Across all configurations, we observe that the performance of PLL models gradually degrades as the proportion of ambiguous samples increases, with (\textit{No PLL}) showing a severe performance drop, highlighting the vulnerability of conventional supervised training on ambiguous ECG annotations.

As shown in Figure~\ref{fig:PTBBaselineperf}, the performance of most PLL methods consistently degrades as $p$ increases. Among the candidate label generation strategies, the \textit{Taxonomy-Driven} ambiguity (Fig.~\ref{fig:PTBBaselineperf}d) is the least challenging, suggesting that this strategy induces weaker forms of ambiguity. In contrast, ambiguity introduced through \textit{Cardiologist-Driven} on \textit{Class-Level} and \textit{Instance-Level} strategies (Figs.~\ref{fig:PTBBaselineperf}b--c) leads to considerable performance declines, which show the difficulty of resolving ambiguities based on cardiologist interpretation. Notably, methods such as DNPL and PRODEN consistently outperform others and demonstrate more stable performance. A similar pattern is observed on the Chapman dataset. Under \textit{Treatment-Driven} ambiguity, most PLL methods maintain strong performance even at higher levels of $p$ (up to $p = 0.8$). For \textit{Model-Driven} ambiguity, approaches like SST and CR show more degradation. This suggests that PLL models relying heavily on structural or semantic priors may struggle when ambiguity arises at the \textit{Instance-Level }rather than from systematic class overlaps.

We further analyze the impact of $p$ and $\epsilon$ in Fig. ~\ref{fig:random_noise}. Specifically, we fix $\epsilon=0.5$ to examine the effect of increasing the proportion of ambiguous samples (Figs. ~\ref{fig:random_noise}a and~\ref{fig:random_noise}c), and fix $p=0.5$ to assess the effect of varying the number of false labels within each instance during training (Figs. ~\ref{fig:random_noise}b and~\ref{fig:random_noise}d). The results show that increasing both parameters leads to performance degradation across all methods, although PLL models experience a noticeably slower decline when $\epsilon$ increases for both datasets. DNPL and PRODEN again demonstrate more robustness to varying levels of ambiguity. Interestingly, most PLL models are more sensitive to the amount of ambiguous samples ($p$) than to the degree of noise within each sample ($\epsilon$). These findings suggest that while existing PLL approaches can tolerate moderate diagnostic ambiguity, their stability strongly depends on the structure and scale of ambiguity. In contrast, models trained without ambiguity-aware mechanisms exhibit sharp performance drops as ambiguity increases.

\subsection{Quantitative Analysis of Ambiguity in Candidate Label Generation Strategies}
\label{sec:ambiguity_analysis}
\begin{figure*}
    \centering
    \begin{subfigure}{0.45\textwidth}
            \centering
            \includegraphics[width=\textwidth]{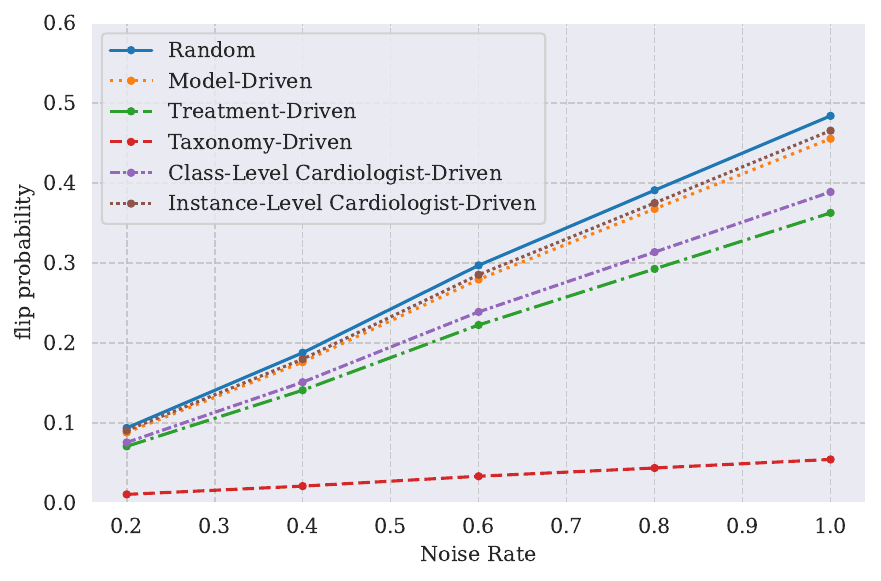}
            \caption{PTB-XL dataset}
    \end{subfigure}
    \begin{subfigure}{0.45\textwidth}
            \centering
            \includegraphics[width=\textwidth]{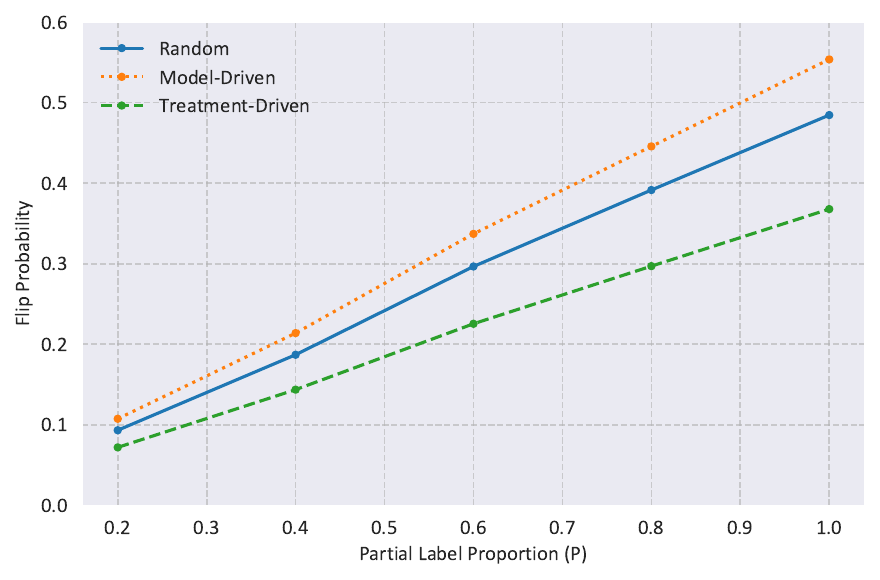}
            \caption{Chapman dataset }
    \end{subfigure}
    \caption{Flip probability across candidate label generation strategies for different proportions of ambiguous samples ($p$) on the PTB-XL and Chapman datasets.}
    \label{fig:ambiguity_rate}
\end{figure*}

We quantitatively examine the degree of ambiguity introduced by different candidate label generation strategies under varying proportions of ambiguous instances ($p$), as illustrated in Fig.~\ref{fig:ambiguity_rate}. To capture this effect, we report the \textit{flip probability}, defined as the average probability of including a false label, computed as the ratio of false positive labels to the total number of negative labels within each instance. As shown in Fig.~\ref{fig:ambiguity_rate}a, the \textit{Random}, \textit{Model-Driven}, and \textit{Instance-Level Cardiologist-Driven} strategies produce higher ambiguity levels compared to the \textit{Taxonomy-Driven} approach, which demonstrates notably lower flip probabilities. This outcome aligns with expectations, since taxonomy-based candidate labels are constrained by class hierarchies and therefore remain closer to the true annotations.

While the \textit{Treatment-Driven} and \textit{Class-Level Cardiologist-Driven} strategies introduce a similar degree of ambiguity, models tend to struggle more with the latter (As observed in Fig. \ref{fig:PTBBaselineperf}). This suggests that beyond the amount of ambiguity, its underlying structure substantially influences how models perform. Finally, the results on the Chapman dataset (Fig.~\ref{fig:ambiguity_rate}b) follow the same pattern observed for PTB-XL, showing consistency between the amount of ambiguity (flip probability) and decrease in models performance (Fig. \ref{fig:ChapmanBaselineperf}). Overall, these findings highlight the importance of considering both the type and the amount of ambiguity when generating candidate labels for a comprehensive evaluation of PLL models.

\subsection{Class-Specific Analysis of PLL Baselines}
\begin{figure}
    \centering
    \includegraphics[width=\linewidth]{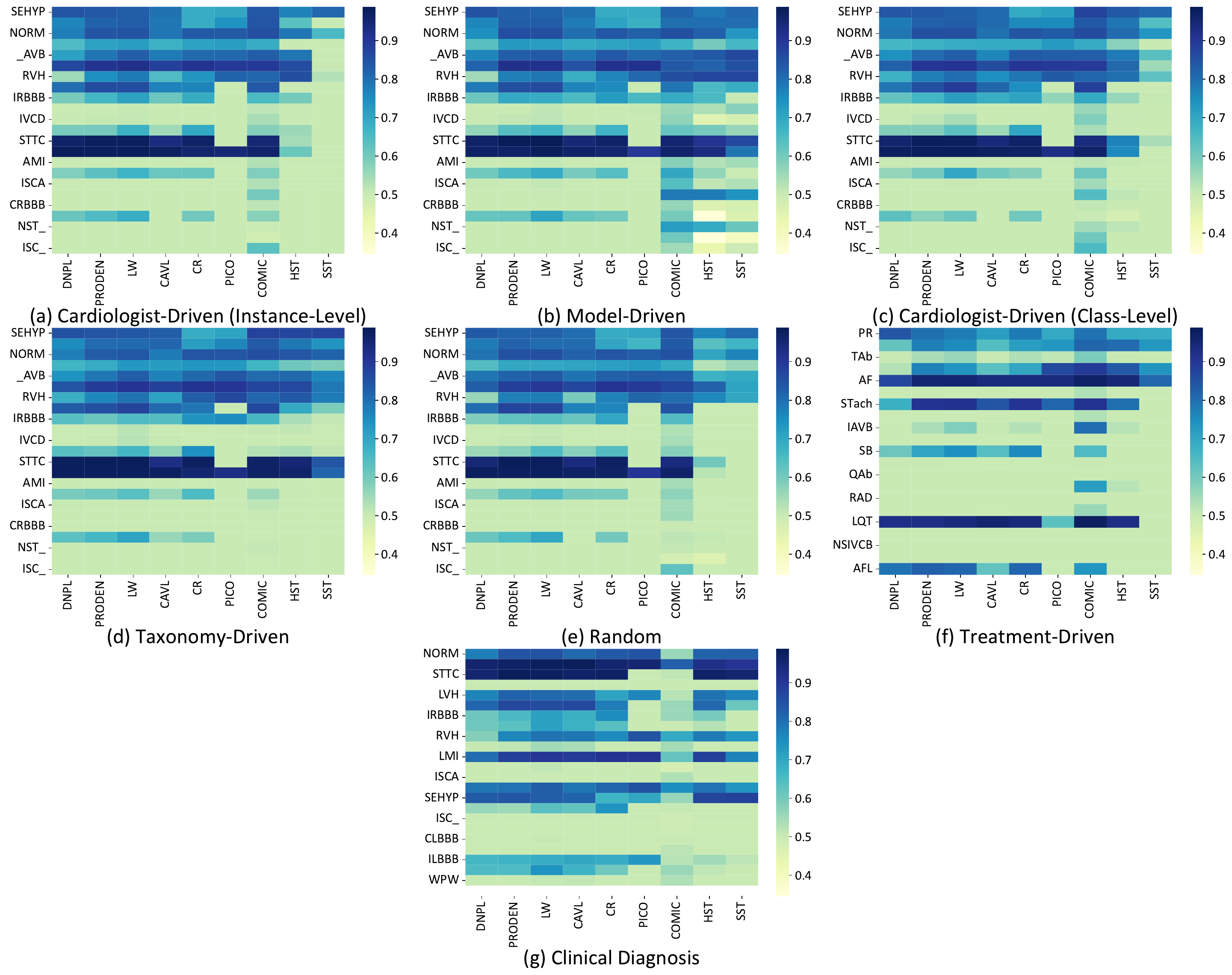}
    \caption{Study of class-specific AUC-ROC performance of PLL methods on PTB-XL dataset.}
    \label{fig:ptb_classperf}
\end{figure}

\begin{figure}
    \centering
    \includegraphics[width=\linewidth]{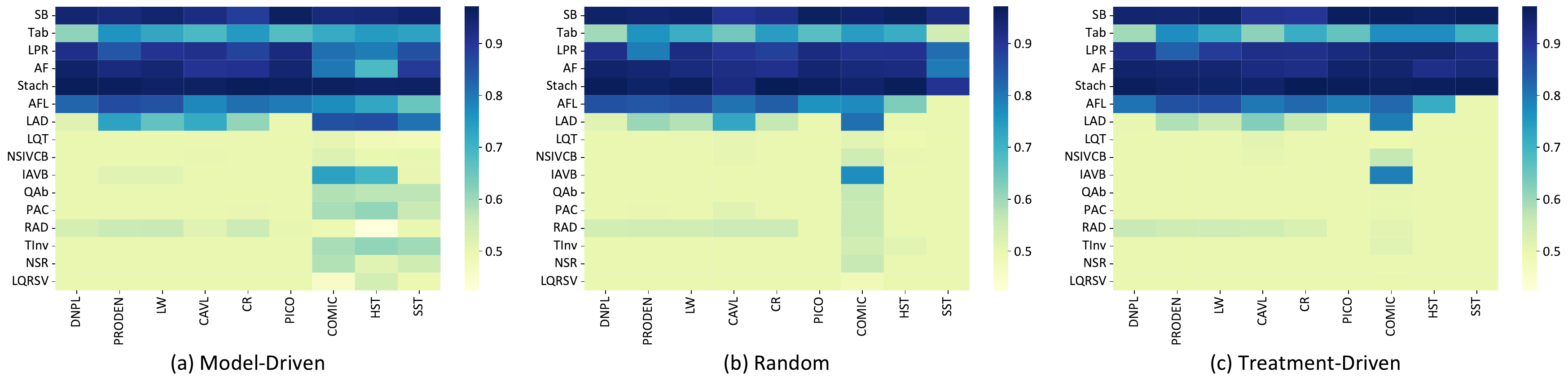}
    \caption{Study of class-specific AUC-ROC performance of PLL methods on Chapman dataset.}
    \label{fig:chapman_classperf}
\end{figure}

To further understand how ambiguity affects different diagnostic categories, we analyze the class-specific AUROC performance of PLL methods on the PTB-XL and Chapman datasets, as illustrated in Figs.~\ref{fig:ptb_classperf} and~\ref{fig:chapman_classperf}. In this analysis, we sort classes by their instance count in the dataset, such that classes with higher count appear toward the top of the heatmaps. The results reveal that the impact of ambiguity is highly class-dependent. Overall, major classes show lower sensitivity to ambiguity, while minor classes experience more pronounced degradation across both datasets. In the PTB-XL dataset, classes such as \textit{IMI} and \textit{STTC} maintain consistently strong performance across ambiguity types, suggesting that their distinctive characteristics make them less prone to ambiguity. In contrast, classes such as \textit{WPW} and \textit{IVCD} show significant performance degradation, despite having relatively higher sample counts than some more robust classes (e.g., \textit{IMI}, \textit{STTC}). This suggests that their diagnostic complexity makes them sensitive to ambiguity. Among the PLL baselines, \textit{COMIC} demonstrates the most consistent performance across the majority of classes, likely due to its ability to handle class-imbalance.

A similar trend is observed for the Chapman dataset, where major classes remain among the most robust, while minor classes are more prone to ambiguity. Once again, \textit{COMIC} maintains more balanced performance across both major and minor classes compared to other baselines. Interestingly, methods originally designed for multi-label settings, such as \textit{HST} and \textit{SST}, display relatively stable behavior across classes under \textit{Model-Driven} ambiguity but decline under other ambiguity types. Collectively, these findings highlight that while models such as \textit{DNPL} and \textit{PRODEN} achieve high overall performance in aggregated metrics (micro F1), their class-level robustness can vary considerably.

\subsection{Limitations and Future Directions}
To our knowledge, CODE Test is currently the only publicly available ECG dataset that systematically records multi-annotator disagreement, making it uniquely suited for studying real clinical ambiguity. While its scale is smaller than large public ECG datasets, the ambiguity it captures reflects genuine inter-clinician uncertainty, a property that cannot be replicated through synthetic ambiguity generation. Expanding the availability of large-scale datasets with authentic ambiguous annotations remains an important direction for future work. 
Although this work provides the first systematic study of adapting PLL methods to ECG diagnosis, several limitations remain. For methods that rely on data augmentation (e.g., contrastive learning or consistency regularization), we restrict our experiments to additive Gaussian noise. Exploring a broader set of ECG-specific augmentations may enhance the performance of these approaches. Additionally, future work could explore a wider variety of encoder backbones to better understand the impact of specific network architectures on performance. Another limitation comes from the standard PLL assumption that the ground-truth label is always included in the candidate set. In real clinical scenarios, annotation errors may exclude the correct label entirely, which is not simulated in the current study. This could be explored in future studies where the correct label could be missing from candidate label sets.  

Our findings highlight several directions for ambiguity-aware learning in ECG analysis. For instance, we observed that sensitivity to hyperparameters limits the generalization of PLL methods in different scenarios. Future work should develop models that are less dependent on hyperparameter choice to ensure generalizability to various ambiguity scenarios in clinical settings. Additionally, many PLL methods show performance degradation on minor classes. Future studies need to account for class imbalance in ambiguous label sets. More specifically, our class-specific analysis highlights the effectiveness of PLL architectures on some classes, while showing the lack of a robust architecture that handles different levels and types of ambiguity experienced in each diagnostic class in real-world clinical disagreement, particularly for classes with higher annotator disagreement. While most of the current PLL frameworks are designed and evaluated with uniform class ambiguities, future work should design ECG-specific PLL frameworks that explicitly account for class-level agreement structure and the varying reliability of clinical annotations across diagnostic categories. Given the annotations by clinicians of varying expertise levels in CODE Test dataset, future work could incorporate annotator reliability into the candidate label construction process and ambiguity-aware model training.

\section{Conclusion}
In this work, we present a comprehensive study of PLL methods for ECG diagnosis under diagnostic ambiguity. We adapt nine PLL algorithms, originally developed for computer vision tasks, to ECG diagnosis and evaluate them across both real-world and controlled ambiguity settings. Specifically, our evaluation uses a real-world dataset of multi-annotator diagnostic disagreements, alongside a range of clinically motivated synthetic ambiguity scenarios, including random, class-dependent (e.g., treatment similarity, disease taxonomy, and cardiologist-driven), and instance-dependent (model- and cardiologist-driven) settings. Our analysis shows that different PLL methods perform differently in dealing with different forms of ambiguity. Specifically, we find that PLL methods outperform classical solutions that do not explicitly consider label ambiguities, with DNPL and PRODEN generally outperforming other PLL approaches. Our class-specific analysis highlights that the type of label ambiguity largely affects the minority classes, which shows the importance of considering class imbalance in designing and evaluating PLL methods for ECG diagnosis.

By leveraging real clinical diagnostic disagreement, this work demonstrates that PLL can effectively learn from ambiguous annotations as they naturally occur in clinical practice. Controlled synthetic experiments confirm and extend this finding, providing insight into which ambiguity structures are most challenging and which PLL design choices confer robustness. At the same time, the results on the CODE Test dataset show that strong performance under synthetic ambiguity does not necessarily translate to real annotation disagreement, highlighting the importance of evaluating ambiguity-aware methods under authentic clinical labeling patterns. Together, these findings provide a foundation for ambiguity-aware ECG learning that is both clinically grounded and systematically evaluated. We identify key limitations of existing PLL approaches and outline several promising directions for future research. We hope this work encourages the development of next-generation PLL frameworks tailored to the challenges of clinical annotation and supports the creation of more robust ECG diagnostic systems.
\begin{acks}
We would like to thank the Natural Sciences and Engineering Research Council of Canada (NSERC) and Ingenuity Labs
Research Institute for partially funding and supporting this work.

\end{acks}

%
\bibliographystyle{ACM-Reference-Format}
\bibliography{IEEEbib}

\end{document}